\documentclass[journal,10pt,twoside]{IEEEtran}
\IEEEoverridecommandlockouts

\usepackage{verbatim}
\usepackage[pdftex]{graphicx}
\usepackage{epsfig,subfigure}
\usepackage{amsmath,amssymb,amsfonts,bm,amscd} 
\usepackage{algpseudocode}
\usepackage{mathtools} 
\usepackage{mathrsfs} 
\usepackage{amsthm} 
\usepackage{setspace}
\usepackage{fancyhdr}
\usepackage{algorithm} 
\usepackage{psfrag}
\usepackage{cite}
\usepackage{makecell}
\usepackage{url}
\usepackage{float}
\usepackage{soul}
\usepackage{color}
\usepackage{multirow}

\usepackage{color, xcolor}
\usepackage{etoolbox}    

\newtheorem{theorem}{Theorem}

\newtheorem{remark}{Remark}

\newtheorem{assumption}{\textbf{Assumption}}
\newtheorem{definition}{\textbf{Definition}}

\newcommand{\rr}{\mathbb{R}}

\newcommand{\Tau}{\mathrm{T}}

\newcommand{\mobility}[0]{ ${V}_{mean}$ ($\text{m/s}$)
                     }

\newcommand{\cruiseconsistency}[0]{${\mathcal{J}}_{mean}$   ($\text{m/s}^3$) 
                     & ${\mathcal{J}}_{max}$          ($\text{m/s}^3$)     
                     & ${\mathcal{P}}_{c}$  ($\text{\%}$)
                    }

\newcommand{\rom}[1]{(\expandafter{\romannumeral #1\relax})}  
\makeatletter
\pretocmd\@bibitem{\color{black}\csname keycolor#1\endcsname}{}{\fail}
\newcommand\citecolor[1]{\@namedef{keycolor#1}{\color{blue}}}
\makeatother

\begin{document}
\title{
 
Barrier-Enhanced Parallel Homotopic Trajectory Optimization for Safety-Critical \\
Autonomous Driving  
 
}  

\author{ 
Lei Zheng, Rui Yang, Michael Yu Wang, \textit{Fellow, IEEE,} and Jun Ma
   \thanks{This work was supported by the National Natural Science Foundation of China under Grant 62303390. \textit{(Corresponding author: Jun Ma.)}}
  \thanks{
    Lei Zheng and Rui Yang are with the Robotics and Autonomous Systems Thrust, The Hong Kong University of Science and Technology (Guangzhou), Guangzhou 511453, China (e-mail: lzheng135@connect.ust.hk; ryang253@connect.hkust-gz.edu.cn).}
    \thanks{Michael Yu Wang is with the School of Engineering, Great Bay University, Dongguan 523000, China (e-mail: mywang@gbu.edu.cn). }
 \thanks{Jun Ma is with the Robotics and Autonomous Systems Thrust, The Hong Kong University of Science and Technology (Guangzhou), Guangzhou 511453, China, and also with the Division of Emerging Interdisciplinary Areas, The Hong Kong University of Science and Technology, Hong Kong SAR, China (e-mail: jun.ma@ust.hk).} 
   \thanks{ {Simulation videos and descriptions of our method are available at \protect\url{https://sites.google.com/view/bphto?pli=1} and  \protect\url{https://youtu.be/ensRDOYWeZ4}}.}
}   
\markboth{IEEE TRANSACTIONS ON INTELLIGENT TRANSPORTATION SYSTEMS, VOL. 26, NO. 2, FEBRUARY 2025}
{Zheng \MakeLowercase{\textit{et al.}}: BPHTO FOR SAFETY-CRITICAL AUTONOMOUS DRIVING} 

	\maketitle  
	 
      \begin{abstract}
    Enforcing safety while preventing overly conservative behaviors is essential for autonomous vehicles to achieve high task performance. 
    In this paper, we propose a barrier-enhanced parallel homotopic trajectory optimization (BPHTO) approach with the over-relaxed alternating direction method of multipliers (ADMM) for real-time integrated decision-making and planning. 
    To facilitate safety interactions between the ego vehicle (EV) and surrounding vehicles, a spatiotemporal safety module exhibiting bi-convexity is developed on the basis of barrier function. 
    Varying barrier coefficients are adopted for different time steps in a planning horizon to account for the motion uncertainties of surrounding HVs and mitigate conservative behaviors.
    Additionally, we exploit the discrete characteristics of driving maneuvers to initialize nominal behavior-oriented free-end homotopic trajectories based on reachability analysis, and each trajectory is locally constrained to a specific driving maneuver while sharing the same task objectives. 
    By leveraging the bi-convexity of the safety module and the kinematics of the EV, we formulate the BPHTO as a bi-convex optimization problem. Then constraint transcription and the over-relaxed ADMM are employed
    to streamline the optimization process, such that multiple trajectories are generated in real time with feasibility guarantees. Through a series of experiments, the proposed development demonstrates improved task accuracy, stability, and consistency in various traffic scenarios using synthetic and real-world traffic datasets.
    \end{abstract}
    
      \noindent Video of the experiments: \protect\url{https://youtu.be/ensRDOYWeZ4} 
	\section{Introduction}
	\label{sec:introd}  
        \IEEEPARstart{E}{nsuring} the safety of autonomous vehicles in dynamic environments is crucial.  The imperative lies in developing a motion planning strategy that ensures safety while keeping high task performance~\cite{claussmann2020review,chen2022milestones,gharavi2023proactive}.  
        This necessitates the strategy to generate safe, feasible, comfortable, and energy-efficient trajectories in real-time replan iterations~\cite{kousik2021safe,paden2016survey,zheng2023spatiotemporal}. However, achieving these characteristics in plans poses substantial challenges. Firstly, challenges stem from the inherently multi-modal nature of motion patterns exhibited by human-driven vehicles (HVs), such as sudden deceleration and cut-in behaviors~\cite{chen2021exploring,hang2023brain,zhou2023interaction}. 
        These challenges pose formidable threats to the safety and driving stability of the autonomous ego vehicle (EV) when interacting with HVs, particularly in dense traffic.  
        Secondly, the non-holonomic kinematic constraints of the EV, coupled with safety constraints, introduce nonlinearity and non-convexity into the planning problem, making it challenging to find feasible trajectories in real time~\cite{ma2022alternating,huang2023decentralized,gharavi2023efficient}. In this context, the optimization landscape features multiple local minima and non-smooth regions, substantially impacting the convergence and performance of gradient-based optimization algorithms~\cite{ghadimi2015optimal}. 
        Thirdly, the swift replanning behavior, especially in selecting the target driving lane, may lead to frequent lane changes, adversely affecting driving safety and consistency. These challenges underscore the crucial requirement for an efficient motion planning framework to tackle the intricacies of autonomous vehicle navigation. 
  
        In general, motion planning for autonomous driving applications can be attempted in a sequential manner. A decision-making module, known as a behavior planner, handles high-level decisions and produces a coarse trajectory, and then a trajectory planner takes these decisions and generates a smooth and feasible trajectory~\cite{schwarting2018planning,sadat2019jointly}. Since the driving maneuvers in the decision-making process are inherently discrete variables (e.g., lane changing and lane keeping), existing behavior planners typically address this challenge by solving the mixed-integer programming (MIP) problem~\cite{qian2016optimal,fabiani2019multi,schwarting2018planning},  which is NP-hard~\cite{nemhauser1988computational}. This complexity becomes particularly critical when the EV needs to operate in multi-lane driving scenarios under dense traffic.
        To tackle these issues, Finite State Machine (FSM)\cite{palatti2021planning,shu2023safety,he2021rule} has been proposed to select the appropriate driving maneuver. Following this, an optimization-based trajectory planner is developed to generate the target trajectory for the EV. Nevertheless, these decoupled motion planning architectures may result in a planned trajectory deviating from the target maneuver, leading to either conservative or aggressive actions~\cite{zhang2021unified}.
  
        Rather than tackling the decision-making and trajectory-planning problems separately, the optimal control framework has been employed to integrate discrete decision variables into a continuous optimization problem~\cite{liu2017path,wang2016optimal,wang2018predictive,ammour2022mpc,tacs2023decision}. In~\cite{liu2017path}, a mixed-integer model predictive control (MPC) scheme with a fail-safe strategy is developed to facilitate the safe interaction of the EV with surrounding vehicles. However, solving this MIP problem poses computational challenges, especially in practical multi-lane autonomous driving scenarios. To streamline the optimization process, this work employs relaxation and constraint enforcement techniques to transform the MIP problem into a nonlinear programming (NLP) problem within a nonlinear MPC (NMPC) framework~\cite{wang2016optimal,wang2018predictive}. In~\cite{zheng2023real}, an optimal control framework integrating behavior and trajectory planning is developed to facilitate the navigation of the EV through a multi-lane dense traffic scenario, leveraging a multi-threading technique. While these studies utilize off-the-shelf solvers to achieve nearly real-time performance,
        it is worth noting that these solvers may struggle to compute a feasible solution due to the nonlinear, non-convex characteristics of the optimization problem. To enhance feasibility, existing interaction-aware MPC techniques either reinitialize the optimal control problem with zero as a starting point~\cite{wang2023interaction} or introduce slack variables when the collision avoidance constraint is not feasible~\cite{zhou2023interaction}.

       To facilitate safety interactions between the EV and surrounding HVs, the reachability analysis~\cite{leung2020infusing} and control barrier function (CBF)~\cite{ames2019control} have been utilized to construct collision avoidance constraints in autonomous driving. 
        In~\cite{brudigam2023stochastic, pek2020fail}, the fail-safe strategy has been designed as a safety filter for the EV based on the obstacle-free reachable set computed through reachability analysis. 
        Although reachability analysis provides formal safety guarantees for the EV, it has the drawback that unsafe regions may expand rapidly over time. Consequently, the planned motions tend to be overly conservative.  
        Alternatively, the CBF with proactive collision avoidance properties is integrated as a safety constraint in the NMPC framework to enhance safety interactions~\cite{zeng2021safety,he2021rule}. However, solving the NMPC becomes computationally burdensome over a long planning horizon (typically exceeding 50 steps) for practical autonomous driving tasks due to the necessity of solving the inverse of the Hessian matrix~\cite{ma2022local}. It is worth noting that these works assume constant speeds for surrounding vehicles, potentially compromising the safety of the EV, especially when surrounding HVs exhibit non-deterministic behaviors, such as abrupt lane changes. 
        
        Considering the uncertain behaviors of surrounding HVs, researchers have extensively implemented partially observable Markov decision process \cite{hubmann2018automated,hubmann2019pomdp, tang2022integrated,li2023pomdp} to address the motion uncertainties of surrounding vehicles. While these works showcase the ability to handle the multi-modal behaviors of surrounding HVs, solving such problems becomes computationally intractable as the problem size increases\cite{li2023marc}. An alternative approach involves employing multiple trajectory optimization methods to handle the multi-modal behaviors of the HVs in a receding horizon planning manner~\cite{chen2022interactive,adajania2022multi,chen2023interactive,de2024topology}. In~\cite{chen2022interactive}, a branch MPC is proposed to optimize over a scenario tree representing possible future behaviors of surrounding uncontrolled agents. Although this method shows promise in facilitating safe interactions between the EV and surrounding HVs, the optimized trajectories tend to be overly conservative, thereby compromising driving efficiency. 
        In \cite{adajania2022multi}, Batch-MPC is proposed for real-time highway autonomous driving through optimizing multiple trajectories based on the alternating minimization algorithm~\cite{tseng1991applications}.
        However, Batch-MPC frequently switches between local optimal trajectories, leading to frequent lane changes and compromising driving consistency. To overcome these significant impediments, a topology-driven planner is developed~\cite{de2024topology}. It iteratively plans multiple evasive trajectories in distinct homotopy classes, ensuring the planner does not change the homotopy class throughout the optimization process.  Additionally, 
        a consistent parameter is introduced in the decision-making module to prevent frequent switching of the homotopy class of the executed trajectory. To further consider the interaction between the EV and HVs, an interactive joint planner (IJP) based on homotopy trajectory optimization is developed~\cite{chen2023interactive}. The IJP simultaneously optimizes multiple free-end homotopy trajectories, each with a distinct endpoint, aiming to explore diverse motions and mitigate the local minima issue in non-convex optimization. However, none of these homotopic trajectory optimization approaches address the safety recovery of the EV, such as the recovery of a safe following distance after being abruptly cut in by other HVs.
        
         In this paper, we present an integrated decision-making and planning scheme for safety-critical autonomous driving with a proposed \textbf{B}arrier-Enhanced \textbf{P}arallel \textbf{H}omotopic \textbf{T}rajectory \textbf{O}ptimization (BPHTO) algorithm. The discrete driving maneuvers of the EV are utilized to construct behavior-oriented free-end homotopic trajectories based on reachability analysis. Subsequently, BPHTO integrates these nominal free-end homotopic trajectories, considering safety and stability, into a bi-convex optimization problem. To ensure feasibility and streamline the optimization process, we employ constraint transcription and the over-relaxed alternating direction method of multipliers (ADMM)~\cite{qian2016optimal} to enable real-time solving of this bi-convex optimization problem.  
         
        The main contributions of this paper are summarized as follows:
	 
        \begin{itemize} 
        \item We propose a BPHTO algorithm to seamlessly integrate decision-making and planning for autonomous driving, which inherently blends discrete maneuver decisions into continuous parallel trajectory optimization. By leveraging reachability analysis, we devise a goal-sampling strategy with warm initialization to determine discrete maneuver homotopy for BPHTO in a receding horizon planning manner. This allows the EV to respond adeptly to surrounding HVs without compromising driving consistency.
        
        \item  We leverage the spatiotemporal information between the EV and HVs to design the spatiotemporal control barrier to enable proactive interaction between the EV and uncertain HVs with safety guarantees. By progressively increasing the barrier coefficient, we effectively account for the motion uncertainties of HVs, enabling the EV to take less conservative actions with safety assurances. 
        Moreover, a rigorous theoretical analysis demonstrates the robustness of safety, showcasing the asymptotic convergence of the EV from an unsafe state to a safe state in the sense of Lyapunov stability.
         
        \item {We exploit the bi-convexity of the kinematics of the EV and the spatiotemporal control barrier to split the BPHTO into several low-dimensional Quadratic Programming (QP) subproblems through over-relaxed ADMM iterations. This strategic approach ensures a feasible solution and enables the EV to execute complex driving tasks in real time.} 
        
        \item We thoroughly demonstrate the improved task performance and safety recovery achieved by our proposed framework through comparative simulations with state-of-the-art algorithms on the intelligent driver model (IDM) and recorded real-world traffic datasets. 
        \end{itemize}
 
        The rest of this paper is structured as follows: The problem statement is introduced in Section~\ref{sec:problem}. 
        Section~\ref{sec:Safety_Constraints} presents the spatiotemporal control barrier for ensuring the safety of the EV. The task-oriented motion for autonomous driving is described in Section~\ref{subsec:task-movement}. We derive a parallelizable optimization scheme BPHTO through over-relaxed ADMM iterations in Section~\ref{subsec:reformulation_PTO}. 
        The validation of the proposed algorithm applied to a safety-critical autonomous vehicle system, using both synthetic and real-world traffic data, is demonstrated in Section~\ref{sec:sim}. A discussion of computational efficiency and driving consistency is presented in Section~\ref{sec:discussion}. Finally, a conclusion is drawn in Section~\ref{sec:con}. 
    \section{Problem Statement}
    \label{sec:problem} 
   
\begin{figure}[tbp]
\centering
\includegraphics[width=0.95\columnwidth]{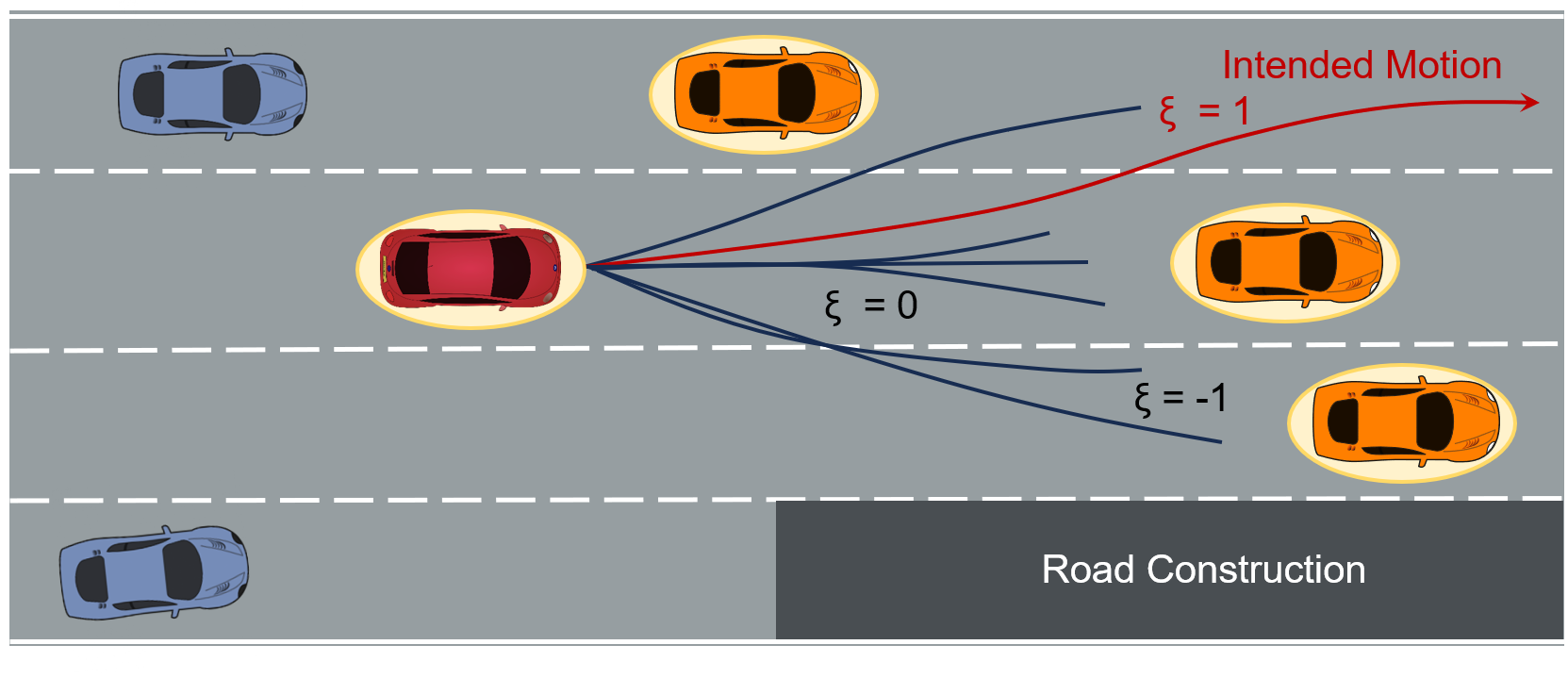}\vspace{-2mm}
    \caption{{ Illustration of the motion of an EV (in red color) in a dynamic cluttered scenario with one lane under road construction ahead. The orange and blue vehicles represent perceived and unperceived HVs, respectively. The EV and the $i$-th HV are represented as ellipse-shaped convex compact set $\mathbb{X}$ and $\mathbb{O}_i $, respectively. The solid red line with an arrow represents the intended trajectory of the EV, while other solid lines denote alternative free-end homotopic candidate trajectories of the EV. Each trajectory shares the same initial state and corresponds to a specific driving maneuver denoted as $\xi$, with values of $\{0, 1, -1\}$, representing lane-keeping, left-lane-change, and right-lane-change behaviors, respectively.}} \vspace{-2mm}
\label{fig:Problem_statement}
\end{figure}

 In this study, we consider multi-lane dense and cluttered driving scenarios, as depicted in Fig.~\ref{fig:Problem_statement}. We adopt Dubin’s car model for the EV, with the yaw rate $\dot{\theta}$ and acceleration $a$ as control inputs~\cite{chen2023interactive}. To facilitate smooth trajectory optimization, we expand the state vector to include control inputs and their derivatives: 
    \begin{equation} \vspace{-1mm}
    \mathbf{x}_k = [p_x\quad p_y \quad \theta \quad \dot{\theta} \quad v \quad a_x\quad a_y\quad j_x\quad j_y ]^T \in \mathcal{X},
       \notag
     \vspace{-1mm} \end{equation}
    where $p_x$ and $p_y$ denote the longitudinal and lateral positions of the EV in the global coordinate, respectively; $v$ denotes the speed of the EV in the global coordinate; $\theta$ represents the heading angle of the EV; $a_x$ and $a_y$ denote the longitudinal and lateral accelerations in the global coordinate, respectively; $j_x$ and $j_y$ denote the longitudinal and lateral jerks in the global coordinate, respectively. 
    In this safety-critical autonomous driving scenario, the autonomous EV encounters substantial challenges when interacting with multiple surrounding HVs that exhibit multi-modal behaviors, such as accelerations, decelerations, and lane changes. These complex interactions compromise driving efficiency and pose significant safety risks. Moreover, the frequent lane changes in this environment can adversely affect the overall driving comfort and jeopardize the safety of the EV. To address this complex problem, we make the following foundational assumptions and definitions:
 \begin{assumption}(\textbf{Safety\ Responsibility}~\cite{shalev2017formal}) 
\label{assumption: Safety_responsibility} When two vehicles are driving in the same direction, if the rear vehicle $c_r$ hits the front vehicle $c_f$ from behind, then the rear vehicle $c_r$ is responsible for the accident.
\end{assumption} 
 \begin{assumption}(\textbf{Perception\ Ability})\label{assumption: communication}
 { An autonomous EV possesses the capability to gather accurate information regarding the current positions and velocities of the $M$ nearest HVs .}
\end{assumption} 
\begin{definition}
(\textbf{Free-end\ Homotopy}~\cite{chen2023interactive}):\
\label{def:homotopy_traj}  Let \(\tau_1: \mathbb{R} \rightarrow \mathcal{X}\) and \(\tau_2: \mathbb{R} \rightarrow \mathcal{X}\) be two continuous trajectories that share the same start point but not necessarily the same endpoint. A continuous mapping \(\Phi: [0, 1] \times \mathbb{R} \rightarrow \mathcal{X}\) is termed a free-end homotopy if it satisfies the following criteria:
\begin{itemize}
  \item \(\Phi(0, \cdot) = \tau_1(\cdot)\),
  \item \(\Phi(1, \cdot) = \tau_2(\cdot)\).
\end{itemize}
If a free-end homotopy \(\Phi\) exists between \(\tau_1\) and \(\tau_2\), then the two trajectories are said to be free-end homotopic, which is a generalization of homotopy as proven in~\cite{chen2023interactive}.
\end{definition} 

We aim to develop an efficient integrated decision-making and planning framework for the EV, enabling safe and stable interactions with HVs while maintaining high task efficiency. 
This framework aims to simultaneously generate multiple free-end homotopic trajectories represented as \begin{equation} \vspace{-1mm}\label{eq:nominal_trajectories}
\mathcal{T}:= \left\{ \mathbf{x}^{(j)}_k \right\}_{j=1}^{N_c},\  k\in\mathcal{I}_0^{N-1},
 \vspace{-1mm} \end{equation}
where $\mathbf{x}_k \in \mathcal{X}\subset \rr^n$ denotes the state vector of the EV at {time instant} $k$;  $N$ and $N_c$ denote the planned time steps and the number of trajectories, respectively. Each free-end homotopic candidate trajectory $\tau_j = \left\{ \mathbf{x}^{(j)}_k \right\}_{k=1}^{N} \in \mathcal{T}  $ is elaborately designed and optimized to represent a specific discrete driving behavior $\xi \in \Xi$, where $\Xi = \{0, 1, -1\}$.
Each behavior is aligned with a target driving lane, as illustrated in Fig.~\ref{fig:Problem_statement}.  

The integrated decision-making and planning framework can be formulated as a finite-constrained NLP problem over a horizon of $N$ steps, as follows:     
    \begin{subequations} 
      \label{problem1}   
      \begin{align} 
       \displaystyle\operatorname*{minimize}_{\substack{
      \Tau}}~~
      &\sum_{j=0}^{N_c-1} \sum_{k=0}^{N-1}  \mathcal{L}(\mathbf{x}^{(j)}_k) +  \phi(\mathbf{x}^{(j)}_N) \label{eq:bp_pto1}\\
 \quad\text { s.t. }~~
          &   \mathbf{x}^{(j)}_0=\mathbf{x}(0), \label{eq:bp_pto2}\\
    &\mathbf{x}^{(j)}_N \in \mathcal{X}^{(j)}_f, \label{eq:bp_pto2}\\
    & \mathbf{x}^{(j)}_{k+1}  = f^{EV}(\mathbf{x}^{(j)}_k),\label{eq:bp_pto3}\\ 
    & \mathbf{O}^{(i)}_{k+1}= \zeta (\mathbf{O}^{(i)}_k), \label{eq:bp_pto4}   \\ 
    &{ \mathbf{dist}(\mathbb{X}^{(j)}_k,\mathbb{O}^{(i)}_k) > d^{(i)}_{\text{safe}}}, \label{eq:bp_pto5}\\
    & \mathbf{x}^{(j)}_k  \in \mathcal{X},	\label{eq:bp_pto6} \\
    &  \forall k\in\mathcal{I}_0^{N-1},\   j\in\mathcal{I}_0^{N_c-1}, \ i\in\mathcal{I}_0^{M-1}.\ \nonumber
    \end{align}  
    \end{subequations} 
    Here, \(\mathcal{I}_0^{N_c-1}\) represents the set of consecutive integers from \(0\) to \(N_c-1\); $\Tau =  \left\{  \tau_1^T,  \tau_2^T,  \ldots, \tau_{N_c}^T\right\}$ represents the optimized trajectories;  $M$ denotes the anticipated number of HVs in trajectory optimization; $\mathbf{x}_0$ denotes the observed initial state vector of the EV;  $f^{EV}$ describes the non-holonomic motion constraints for the EV, ensuring that the EV's motion adheres to its physical limitations.
     Each $\mathbf{O}^{(i)}$ denotes the state vector of the $i$-th HV in the environment, and $\zeta$ characterizes the predicted motion for these HVs.  The term $\mathcal{L}$ represents the running cost, designed to encode specific task requirements, such as smoothness and task accuracy, and $\phi$ is the terminal cost function for stabilization consideration. 
    \begin{remark} 
     The target state set $\mathcal{X}^{(j)}_f$ is designed to optimize the $j$-th candidate trajectory to a desired driving lane, with the attendant outcome of a driving behavior $\xi$.
    \end{remark}
 
    \begin{remark} 
    The optimization problem \eqref{eq:bp_pto1}--\eqref{eq:bp_pto6} yields meticulously optimized and feasible free-end homotopic trajectories, each of which is associated with a specific driving behavior. It effectively blends discrete maneuver decisions with continuous trajectory generation for motion planning.
    \end{remark}
  
    The challenges of designing and optimizing this NLP problem are fourfold:  
 \begin{itemize} 
    \item [(1)] 
    \emph{Safety:} The motion of HVs exhibits multi-modal driving behaviors that are difficult to predict accurately.  This challenge makes it difficult to strictly satisfy safety constraints~(\ref{eq:bp_pto5}) over a long planning horizon,
    \item [(2)] \emph{Feasibility}: The non-holonomic constraint~(\ref{eq:bp_pto4}) and safety constraint~(\ref{eq:bp_pto5}) are typically nonlinear and non-convex, posing significant challenges for finding feasible solutions.
    \item [(3)] 
    \emph{Computational Efficiency:} The proposed approach must efficiently optimize multiple nominal free-end homotopic trajectories in real time during interactions with multiple HVs, ensuring prompt decision-making and adaptability to dynamic environments.
   \item [(4)] \emph{Safety-Performance Trade-off:} Balancing multiple intricate objectives and constraints, such as safety, motion stability, task accuracy, and control limits, is essential in designing an NLP problem for the EV. The presence of uncertainties in the motion of surrounding HVs adds further complexity to the problem, making it more challenging to achieve the desired task performance while avoiding collisions, frequent lane changes, and overly cautious behaviors.   
\end{itemize}   

\section{Spatiotemporal Control Barrier} 
\label{sec:Safety_Constraints} 
\subsection{Trajectory Parameterization}
\label{subsec:Trajectory_para} 
In this study, each trajectory is optimized over a compact time interval with a finite duration of $T$. 
To obtain optimal and smoothly controllable homotopic trajectories for the EV, we employ a representation based on  B\'ezier curves~\cite{farouki2012bernstein,tong2023}, facilitating continuous differentiable and optimized trajectories for the EV.  
For an $m$ dimensional and $n$-th order  B\'ezier curve, the representation is given by:
\begin{equation} \vspace{-1mm}
\label{eq:continuous_traj}
\quad\mathbf{C}^{(j)}(\nu) = \sum_{i=0}^{n}  {B}_{i,n} (\nu)\mathbf{P}^{(j)}_{i},  j\in\mathcal{I}_0^{N_c-1},
 \vspace{-1mm} \end{equation} 
where  $\mathbf{P}^{(j)}_{i}  \in \rr^m$ represents control points or  B\'ezier coefficients to be optimized for the $j$-th trajectory. Specifically, we define $\mathbf{P}^{(j)}_{i} = [c^{(j)}_{x,i}\quad c^{(j)}_{y,i}\quad c^{(j)}_{\theta,i}]^T$, where $c^{(j)}_{x,i}$, $c^{(j)}_{y,i}$, and $c^{(j)}_{\theta,i}$ denote the coefficients for the longitudinal position $p_x$, lateral position $p_y$, and heading angle $\theta$, respectively. The Bernstein polynomial basis ${B}_{i,n}$ is defined as
\begin{equation} \vspace{-1mm}
{B}_{i,n} (\nu)  =  \binom{n}{i} \nu^i (1 - \nu)^{n - i},
 \vspace{-1mm} \end{equation}  
where $\nu = \frac{t - t_0}{T} \in [0, 1]$ is the parameter varying from $0$ to $1$. Here, $t_0$  represents the initial time, $t = t_0 + k\delta t$ is the current {time instant} of the trajectory,  $k$ denotes discrete time steps, and $\delta t = T/N$ is the corresponding discrete time interval. 

As a result, the trajectory sequences can be expressed as follows:
\begin{equation} \vspace{-1mm}
\label{eq:discrete_traj}
\left\{ \mathbf{C}^{(j)}_k \right\}_{k=0}^{N-1} = \mathbf{W}^T_{P,j} \mathbf{W}_B, j\in\mathcal{I}_0^{N_c-1},
 \vspace{-1mm} \end{equation} 
{where $ \mathbf{C}^{(j)}_k = [p^{(j)}_{x,k}\quad p^{(j)}_{y,k}\quad \theta^{(j)}_{k}]^T $ represents the longitudinal position, lateral position, and heading angle for the $j$-th trajectory at {time instant} \(k\);}  $\mathbf{W}_{P,j} = [\mathbf{P}^{(j)}_0\quad  \mathbf{P}^{(j)}_1 \quad \dots  \quad\mathbf{P}^{(j)}_n]^T \in \rr^{(n+1)\times 3}$ denotes the matrix of control points to be optimized; $ \mathbf{W}_B = [\mathbf{B}_{0}\quad \mathbf{B}_{1}\quad \dots \quad  \mathbf{B}_{n}]^T \in \rr^{(n+1)\times N}$ is a constant basis matrix, where $\mathbf{B}_{i} = [B_{i,n}(\delta t)\quad B_{i,n}(2\delta t)\quad \cdots\quad B_{i,n}(N\delta t)]^T \in \rr^N$. We can derive this $C^1$ continuity trajectory~(\ref{eq:continuous_traj}) to get its velocity, acceleration, and jerk profiles. Note that the discrete integrator model inherent in high-order ($n\geq 4$)  B\'ezier polynomial trajectories~(\ref{eq:continuous_traj}) constrains the motion of the EV.
 
\subsection{Spatiotemporal Control Barrier} \label{subsec:Safety_Constraints} 
To achieve high-performance safe driving, the EV must navigate to its goal state while avoiding collisions with nearby HVs that exhibit multi-modal behaviors. This necessitates the consideration of spatiotemporal relative positions and angles between the EV and surrounding HVs. While the safety constraint~(\ref{eq:bp_pto5}) provides a sufficient but unnecessary condition for collision avoidance, strictly adhering to these constraints can typically result in overly cautious driving maneuvers, comprising task performance. 
In this subsection, we develop an efficient spatiotemporal control barrier for the EV. This spatiotemporal control barrier enables the EV to safely interact with uncertain surrounding HVs while avoiding overly cautious driving behaviors. Additionally, a rigorous robustness safety analysis is provided. 
\subsubsection{Safety Representations}
\label{subsec:CBF} 
 The unsafe set, safe set, and interior safe set for the EV are defined as follows: 
\begin{subequations} {\vspace{-1mm} \begin{align}  Out(\mathcal{S}) &:= \{\mathbf{x} \in \mathcal{X} \mid h(\mathbf{x}, \mathbf{O}^{(i)}) < 0,\  \forall i\in\mathcal{I}_0^{M-1}\},  \label{eq:unsafe} \\ 
\mathcal{S} &:=\{\mathbf{x}\in\mathcal{X} \mid h(\mathbf{x}, \mathbf{O}^{(i)})\geq0,\  \forall i\in\mathcal{I}_0^{M-1}\}, \label{eq:safe} \\
Int(\mathcal{S}) &:= \{\mathbf{x} \in \mathcal{X} \mid h(\mathbf{x}, \mathbf{O}^{(i)}) > 0,\  \forall i\in\mathcal{I}_0^{M-1}\}, \label{eq:int_safe}   \end{align}} \vspace{-1mm} \end{subequations} 
{where $h$ is a discrete-time barrier function to facilitate safety interactions through the following definition:}
{\begin{definition}(\cite{zeng2021enhancing})
\label{def:discrete_cbf}
 A function $h$ is said to be a discrete-time  barrier function (BF) with respect to the set $\mathcal{S}$ \eqref{eq:unsafe}--\eqref{eq:int_safe}, if there exists a barrier coefficient $\alpha_k   \in (0, 1)$  with $\mathcal{S} \subset  \mathcal{X}$, such that
 \begin{equation} \vspace{-1mm}\label{eq:discrete_cbf}
 \Delta  h(\mathbf{x}_k, \mathbf{O}^{(i)}_{k} ) +\alpha_k h(
 \mathbf{x}_{k-1}, \mathbf{O}^{(i)}_{k-1}  )> 0,
 \vspace{-1mm} \end{equation}
where $ \Delta  h(\mathbf{x}_k, \mathbf{O}^{(i)}_{k} )  :=   h(\mathbf{x}_k, \mathbf{O}^{(i)}_{k} ) -   h(\mathbf{x}_{k-1}, \mathbf{O}^{(i)}_{k-1} )$.
\end{definition}   }
\begin{definition}(\textbf{Forward Invariably Safe Set})\label{def:safety_interaction}  { Given an initial safe state \(\mathbf{x}_0 \in \mathcal{S}\), a barrier function \(h\), and the future trajectories of the \(M\) nearest HVs \(\left\{\mathbf{O}^{(i)}_k\right\}_{k=0}^{T}, \forall i \in \mathcal{I}_0^{M-1}\), the set \(\mathcal{S}\) is said to be a \textit{Forward Invariably Safe Set} if, for all \(k > 0\), the following condition holds:}
\begin{equation}
  {h(\mathbf{x}_k, \mathbf{O}^{(i)}_k) \geq 0, \quad \forall i \in \mathcal{I}_0^{M-1}. }  
\end{equation}
\end{definition}

   \subsubsection{Invariably Safety Constraints} 
 Following the approach outlined in \cite{rastgar2020novel,adajania2022multi}, we leverage the polar representation of Euclidean distance between the EV and surrounding HVs to obtain the following safety constraints:
\begin{eqnarray}
\left\{
    \begin{aligned}
      p_{x,k} &= o_{x,k}^{(i)} + l_x^{(i)}  d^{(i)}_k  \cos (\omega^{(i)}_k), \\
      p_{y,k} &= o^{(i)}_{y,k} + l_y^{(i)}  d^{(i)}_k \sin (\omega^{(i)}_k),  \\
      d^{(i)}_k & \geq 1, \forall  i\in\mathcal{I}_0^{M-1},
    \end{aligned}
\right.
\label{eq:polar_safety1}
\end{eqnarray} 
where $p_{x,k}$ and $ p_{y,k}$ represent the longitudinal and lateral positions
of the EV at {time instant $k$, respectively; $ o_{x,k}^{(i)}$ and $ o_{x,k}^{(i)}$ denote the longitudinal and lateral positions of the $i$-th surrounding HV at time instant $k$, respectively; $l_x^{(i)}$ and $l_y^{(i)}$ denote the length of major and minor axes of the safe ellipse.
The variable $\omega^{(i)}_k \in [0, \pi]$ denotes the angle of safe ellipse between the EV and $i$-th surrounding HV at time instant $k$.}

Note that the variable \(d^{(i)}_k\) functions as a scaling factor influencing the size of the safety region associated with the \(i\)-th surrounding HV at {time instant} \(k\). Larger values of \(d^{(i)}_k\) correspond to larger safety regions, promoting increased separation, while a value close to 1 indicates a more compact safety region, maintaining a non-zero separation distance for safety considerations. 
 
Referring to \cite{adajania2023amswarm}, we can further formulate barrier function $h$ with respect to the safety constraint~(\ref{eq:polar_safety1}) as
     \begin{equation} \vspace{-1mm}
  h(\mathbf{x}_k, \mathbf{O}^{(i)}_k) =  d^{(i)}_k -1.
\label{eq:barrier_function}
 \vspace{-1mm} \end{equation}  

Hence, a BF constraint can be formulated to facilitate proactively collision avoidance as 
{\begin{equation} \vspace{-1mm}
\label{eq:spatiotemporal_barrier_cons}
\Delta  h(\mathbf{x}_k, \mathbf{O}^{(i)}_k) + \alpha_k (h(\mathbf{x}_{k-1}, \mathbf{O}^{(i)}_{k-1})  > 0 ,
 \vspace{-1mm} \end{equation} } 
which can be explicitly expressed as
   \begin{equation} \vspace{-1mm} 
\label{eq:barrier_cons_polar}
 d^{(i)}_{k} -1  - (1- \alpha_k )  ( d^{(i)}_{k-1} -1) > 0. 
  \vspace{-1mm} \end{equation}  

 The constraint~\eqref{eq:spatiotemporal_barrier_cons} ensures that the EV with an initial safe state $\mathbf{x}_0 \in \mathcal{S}$ remains within the forward invariably safe set  $\mathcal{S}$, as elaborated in~\cite[Proposition 4]{agrawal2017discrete}. 
 
 As a result, we can transform the original safety constraint \eqref{eq:polar_safety1} into the following spatiotemporal control barrier safety constraint:
\begin{eqnarray}
\left\{
    \begin{aligned}
      p_{x,k} &= o_{x,k}^{(i)} + l_x^{(i)}  d^{(i)}_k  \cos (\omega^{(i)}_k), \\
      p_{y,k} &= o^{(i)}_{y,k} + l_y^{(i)}  d^{(i)}_k \sin (\omega^{(i)}_k),  \\
   \Delta  h&{(\mathbf{x}_k, \mathbf{O}^{(i)}_k) + \alpha_k  h(\mathbf{x}_{k-1}, \mathbf{O}^{(i)}_{k-1})  > 0} ,   \forall  i\in\mathcal{I}_0^{M-1}.   
    \end{aligned}
\right.
\label{eq:polar_safety2}
\end{eqnarray} 

We can further derive the closed-form value of $\omega^{(i)}_k$ as follows:
\begin{equation} \vspace{-1mm}
    \omega^{(i)}_k = \arctan\left(l_x^{(i)}( p_{y,k}  - o_{y,k}^{(i)}), \, l_y^{(i)}( p_{x,k}  - o_{x,k}^{(i)})\right).
    \label{eq:angle_value}
 \vspace{-1mm} \end{equation}

\begin{theorem}
\label{theorem:1}
Let $h$ be a discrete-time BF for the EV under Assumptions 1-2. Then, the EV, starting from an initial state $\mathbf{x}_{k-1} \in Int(\mathcal{S})$, can proactively avoid collisions with surrounding HVs with guaranteed safety if the constraint \eqref{eq:spatiotemporal_barrier_cons} is satisfied.
\end{theorem} 
\begin{IEEEproof}
Given an initial state $\mathbf{x}_{k-1} \in Int(\mathcal{S})$, we can derive
the barrier function $ h(\mathbf{x}_{k-1}, \mathbf{O}^{(i)}_{k-1}) > 0 $.  

{With Definition \ref{def:discrete_cbf}}, we can express \eqref{eq:spatiotemporal_barrier_cons} as:
\begin{align} \vspace{-1mm}
\begin{aligned}
 & h(\mathbf{x}_{k}, \mathbf{O}^{(i)}_{k}) >  (1-\alpha_k)  h(\mathbf{x}_{k-1}, \mathbf{O}^{(i)}_{k-1}),
\end{aligned}
\vspace{-1mm} \end{align}
where $\alpha_k \in (0, 1), \forall h \neq 0$. 
This leads to the following result:
\begin{equation} \vspace{-1mm}
h(\mathbf{x}_{k}, \mathbf{O}^{(i)}_{k}) > 0.
 \vspace{-1mm} \end{equation}
As a result, the EV, starting from an initial safe state, remains within the safety set $\mathcal{S}$. 

Moreover, the parameter $\alpha_k \in \rr^+$ serves as a barrier coefficient that influences the changing rate of the safety function $h$ during planning. If $\alpha_k = 1$, the safety constraint (\ref{eq:polar_safety2}) reverts to its original form (\ref{eq:polar_safety1}). Consequently, the constraint does not confine optimization until near safety violation ($h=0$). In contrast, smaller values of \(\alpha_k\) contribute to a more stable adjustment of the safety barrier, promoting proactive collision avoidance, as shown in~\cite{zeng2021safety,adajania2023amswarm}. Conversely, larger values allow for more aggressive maneuvers and less conservative driving behaviors.  
This completes the proof of Theorem 1.
\end{IEEEproof}  

\begin{remark} 
\label{remark:5}
{In dense traffic, accurately predicting the motion of surrounding HVs is challenging, particularly in subsequent planning steps. To tackle this challenge, we adopt a strategy where the parameter \(\alpha_k\) gradually increases throughout the planning horizon \(N\). This approach strikes a balance between task performance and safety. As \(\alpha_k\) grows, it expands the feasible state space and diminishes the influence of safety constraints on the NLP problem \eqref{problem1} over time, as outlined in~\cite{zeng2021safety}. Consequently, the planner can prioritize the driving task in the short term while ensuring the EV's safety in the long term, leading to less conservative actions.}
\end{remark} 
\subsubsection{Robustness of Safety} 
In certain situations, such as abrupt lane changes by surrounding HVs with nonstationary dynamics, safety constraints (\ref{eq:polar_safety2})  may be violated. However, this violation does not necessarily result in a collision with the EV since the axis lengths of the safe ellipse typically exceed the specified collision size limit. 
In this subsection, we further investigate the robustness of the spatiotemporal control barrier safety constraints (\ref{eq:polar_safety2}) regarding safety recovery. The analysis is conducted with an initial unsafe state $\mathbf{x_0} \in Out(\mathcal{S})$, aiming to guide the EV to asymptotically converge to the safe set $\mathcal{S}$ from an unsafe state $Out(\mathcal{S})$. 
\begin{theorem}
\label{theorem:2}
Let $h$ be a discrete-time BF for the EV under Assumptions 1-2. Then, the unsafe state $\mathbf{x}_{k-1} \in Out(\mathcal{S})$ asymptotically converges to the forward invariably safe set $\mathcal{S}$ if the following constraint is satisfied: 
\begin{align} \vspace{-1mm}
\label{eq:barrier_cons}
\begin{aligned}
\Delta  h&(\mathbf{x}_k, \mathbf{O}^{(i)}_k) + {\alpha_k  h(\mathbf{x}_{k-1}, \mathbf{O}^{(i)}_{k-1})}  \geq 0.
\end{aligned}
\vspace{-1mm} \end{align} 
\end{theorem} 
\begin{IEEEproof}
With the barrier function $h$, we define a positive definite Lyapunov function $V:\mathbb{R}^{n}\rightarrow\mathbb{R}$ as follows:
{\begin{equation} \vspace{-1mm}
V(\mathbf{x}_k, \mathbf{O}^{(i)}_k)   =
\begin{cases}
0 & \text{if } \mathbf{x}_k \in \mathcal{S}, \\
 | h(\mathbf{x}_k, \mathbf{O}^{(i)}_k)  |^2 & \text{if } \mathbf{x}_k \in Out(\mathcal{S}). 
\end{cases}
\label{eq:Lyapunov_func}
 \vspace{-1mm} \end{equation}
Based on the constraint~(\ref{eq:barrier_cons}), we can derive the following inequality:
\begin{alignat}{2} \vspace{-1mm}
       | h(\mathbf{x}_k, \mathbf{O}^{(i)}_k) |^2 &\leq | h(\mathbf{x}_{k-1}, \mathbf{O}^{(i)}_{k-1}) - \alpha_k  h(\mathbf{x}_{k-1}, \mathbf{O}^{(i)}_{k-1}) |^2 \nonumber \\
    &= (1 - \alpha_k)^2 | h(\mathbf{x}_{k-1}, \mathbf{O}^{(i)}_{k-1})  |^2, \label{eq:sub_qp1} 
\vspace{-1mm} \end{alignat}
where $h  < 0$. }


{Utilizing the properties of the Lyapunov function, we can derive:
\begin{align} \vspace{-1mm}
\begin{aligned}
\Delta &V(\mathbf{x}_k, \mathbf{O}^{(i)}_k) = V(\mathbf{x}_k, \mathbf{O}^{(i)}_k) -  V(\mathbf{x}_{k-1}, \mathbf{O}^{(i)}_{k-1}) \\
&=  | h(\mathbf{x}_k, \mathbf{O}^{(i)}_k)  |^2  -   | h( \mathbf{x}_{k-1}, \mathbf{O}^{(i)}_{k-1}) |^2  \\
&\leq (1- \alpha_k)^2 |  h(\mathbf{x}_{k-1}, \mathbf{O}^{(i)}_{k-1})  |^2 -  | h(\mathbf{x}_{k-1}, \mathbf{O}^{(i)}_{k-1}) |^2 \\
&\leq  ( ( 1- \alpha_k )^2 - 1 ) |  h( \mathbf{x}_{k-1}, \mathbf{O}^{(i)}_{k-1})  |^2\\
&= ( ( 1- \alpha_k )^2 - 1 ) V(\mathbf{x}_{k-1}, \mathbf{O}^{(i)}_{k-1}).  
\end{aligned}
\vspace{-1mm} \end{align}}

{Consequently, the following inequality holds:
\begin{equation} \vspace{-1mm}
\Delta V(\mathbf{x}_k, \mathbf{O}^{(i)}_k) \leq -c_kV(\mathbf{x}_{k-1}, \mathbf{O}^{(i)}_{k-1}),
 \vspace{-1mm} \end{equation}
where $c_k = (1 - ( 1- \alpha_k )^2 ) \in (0,1)$. }

If there exists no solution that can stay identically in $\mathcal{S}$, other than the trivial solution $ x_k \in \mathcal{S}$, then the origin is asymptotically stable. This result indicates that the state of the EV with $h < 0$ will asymptotically converge to the safe set $\mathcal{S}$ in the sense of Lyapunov stability. This completes the proof of Theorem 2.
\end{IEEEproof}   
Note that the parameter $\alpha_k\in\mathbb{R^{+}}$ adjusts the aggressiveness of safety recovery. We adopt a progressively increasing $\alpha_k\in\mathbb{R^{+}}$ along the planning horizon, allowing for a stable safety recovery.  

\section{Task-Oriented and Maneuver-oriented Motion}
\label{subsec:task-movement}
In this study, each maneuver is aligned with a candidate trajectory directed towards a specific driving lane, resulting in maneuver homotopy, as illustrated in Fig.~\ref{fig:Problem_statement}. The optimization process involves sampling target points  $ \left\{ \mathbf{x}^{(j)}_g \right\}_{j=1}^{N_c}$  within the current and neighboring lanes. These points guide trajectory optimization to remedy the local minimum issue in the non-convex motion space, considering dynamic feasibility and driving stability. 
    \subsection{Dynamic Feasibility} 
    The engine force of the ego vehicle limits the acceleration and braking as follows:  
    \begin{subequations} \vspace{-1mm}
    \begin{align}  
         a_{x, \min}&\leq a^{(j)}_{x,k} \leq a_{x, \max},\forall k\in\mathcal{I}_0^{N-1}, j\in\mathcal{I}_0^{N_c-1},  
         \label{eq:longitudinal_acc_cons} \\
         a_{y, \min}&\leq a^{(j)}_{y,k} \leq a_{y, \max}, \forall k\in\mathcal{I}_0^{N-1}, j\in\mathcal{I}_0^{N_c-1},
             \label{eq:lateral_acc_cons}
       \end{align}  \end{subequations} where $a_{x, \max}$ and $a_{y, \max}$ denote the  maximum longitudinal and lateral acceleration, respectively; $a_{x, \min}$ and $a_{y, \min}$ denote the minimum longitudinal and lateral deceleration, respectively.   
       The motion of the EV is further constrained by nonholonomic constraints:
    \begin{eqnarray}
    \left\{
        \begin{aligned}
          \dot{p}^{(j)}_{x,k} -& v^{(j)}_k \cos (\theta^{(j)}_k) = 0, \forall  k\in\mathcal{I}_0^{N-1}, j\in\mathcal{I}_0^{N_c-1},\\
           \dot{p}^{(j)}_{y,k}- &v^{(j)}_k\sin (\theta^{(j)}_k) = 0,  \forall  k\in\mathcal{I}_0^{N-1}, j\in\mathcal{I}_0^{N_c-1}.
        \end{aligned}
    \right.
    \label{eq:nonholonomic_cons}
    \end{eqnarray}    

  Following the approach outlined in \cite{rastgar2020novel,adajania2023amswarm}, we can further derive the following constraint of $\theta^{(j)}_k$ and closed-form solution of $v^{(j)}_k$ :
      \begin{equation} \vspace{-1mm}
                 \theta^{(j)}_k - \arctan \left(\frac{\dot{p}^{(j)}_{y,k}}{\dot{p}^{(j)}_{x,k}}\right) = 0,  \forall  k\in\mathcal{I}_0^{N-1}, j\in\mathcal{I}_0^{N_c-1},
                 \label{eq:polar_nonholonomic_cons} 
         \vspace{-1mm} \end{equation}
      \begin{equation} \vspace{-1mm}
                 v^{(j)}_k = \sqrt{(\dot{p}^{(j)}_{x,k} )^2 +  (\dot{p}^{(j)}_{y,k} )^2},  \forall  k\in\mathcal{I}_0^{N-1}, j\in\mathcal{I}_0^{N_c-1}.
                 \label{eq:polar_vel} 
         \vspace{-1mm} \end{equation}
    \subsection{Task-Oriented Motion} 
    To generate parallel free-end homotopic trajectories,  we enforce the final longitudinal position ${p}^{(j)}_{x,N}$ and lateral position ${p}^{(j)}_{y,N}$ of each trajectory to align with target sampling points. To achieve this, we introduce the following state constraints to guide the optimized trajectory to the desired state with continuity consideration:  
\begin{subequations} \vspace{-1mm} 
\begin{align}  
  {p}^{(j)}_{x,0} & = {p}_{x}(0),\quad {p}^{(j)}_{y,0} = {p}_{y}(0), \label{eq:init_pos} \\ 
    {v}^{(j)}_{x,0} & = {v}(0) \cos(\theta(0)),\quad {v}^{(j)}_{y,0} = {v}(0)\sin(\theta(0)),  \label{eq:init_vel} \\ 
  {p}^{(j)}_{x,N}  & =   {p}^{(j)}_{x,g},\quad {p}^{(j)}_{y,N}=  {p}^{(j)}_{y,g}, \label{eq:final_pos} 
\end{align}   
\end{subequations} 
 where {$j\in\mathcal{I}_0^{N_c-1}$; $v(0)$ and $\theta(0) $ represent the initial velocity and heading angle of the EV, respectively;} ${p}_{x}(0)$ and ${p}_{y}(0)$ represent the initial position of the EV. The target position sampling mechanism to determine the target longitudinal position $ {p}^{(j)}_{x,g}$ and lateral position $ {p}^{(j)}_{y,g}$ of the $j$-th candidate trajectory will be elaborated in Section \ref{subsubsec:goal_points}.

To ensure a stable driving mode during the generation of each maneuver homotopy, we enforce the planned trajectory to exhibit a tiny heading angle and yaw rate, as specified  by the following terminal driving stability constraints:  
\begin{subequations} \vspace{-1mm} 
\begin{align}  
    {\theta}^{(j)}_0 &=  \theta(0), \quad\dot{\theta}^{(j)}_0 =  \dot{\theta}(0), \label{eq:init_theta} \\ 
    {\theta}^{(j)}_N &= 0, \quad\dot{\theta}^{(j)}_N = 0,  \label{eq:final_theta} 
\end{align}  
\end{subequations} for all $j\in\mathcal{I}_0^{N_c-1}$,
where $\dot{\theta}(0) $ represents the initial yaw rate of the EV. 
Furthermore, to achieve a comfortable driving experience, the jerk needs to be constrained as follows:
\begin{subequations} \vspace{-1mm}
\begin{align} 
     j_{x, \min}\leq j^{(j)}_{x,k} \leq j_{x, \max}, \forall k\in\mathcal{I}_0^{N-1}, j\in\mathcal{I}_0^{N_c-1}, 
     \label{eq:longitudinal_jerk_cons} \\
     j_{y, \min}\leq j^{(j)}_{y,k} \leq j_{y, \max}, \forall k\in\mathcal{I}_0^{N-1}, j\in\mathcal{I}_0^{N_c-1},
         \label{eq:lateral_jerk_cons}  
 \end{align}
 \end{subequations}
where $j_{x, \max}$ and $j_{x, \min}$ denote the maximum and minimum allowable longitudinal acceleration, respectively; $j_{y, \max}$ and $j_{y, \min}$ denote the maximum and minimum allowable lateral deceleration, respectively. The comfortable longitudinal and lateral jerk values have been comprehensively discussed in~\cite{bae2020self}. 

\subsection{Sampling Points Generation} 
\label{subsubsec:goal_points}
To enhance the quality of free-end homotopic trajectory optimization, the target position in \eqref{eq:final_pos} for each trajectory should account for not only the current state vector of the EV but also the state limits of the EV and the future motion of surrounding HVs.

In this study, we develop a warm initialization strategy to determine maneuver homotopy in a receding horizon planning manner to facilitate driving consistency. Specifically, in each replanning step, the lateral goal position ${p}^{(j)}_{y,g},\forall j\in\mathcal{I}_0^{N_c-1}$, for each nominal free-end homotopic trajectory, evolves based on the last optimal trajectory and maneuver as follows: 
\begin{equation}\vspace{-1mm}
     \mathbf{P}_{y,g} =  {p}^*_{y,g} + \delta \mathbf{y}.
     \label{eq:lateral_goal}
 \vspace{-1mm}  \end{equation}  
Here, $ \mathbf{P}_{y,g} =  \left\{ p^{(j)}_{y,g} \right\}_{j=1}^{N_c} \in \mathbb{R}^{1\times N_c}$ represents the target lateral goal vector for $N_c$ nominal free-end homotopic trajectories. ${p}^*_{y,g}$ denotes the lateral position of the optimal trajectory $\xi^*$ chosen from the set of last candidate trajectories, as elaborated in Section \ref{subsec:decision_making}. The maneuver adjustment vector $\delta \mathbf{y} \in \mathbb{R}^{1\times N_c}$ is typically designed as a symmetrical vector centered around zero. It includes positive, negative, and zero values to ensure alignment with the latest optimal maneuver, maintaining consistency and reliability in adjustments.
For example, in Fig.~\ref{fig:Problem_statement}, a mean value of zero indicates maintaining the last decision maneuver ($\xi = 0$), negative values imply right lane changing ($\xi = -1$), generating trajectories below the current lane, while positive values denote left lane changing ($\xi = 1$), generating trajectories above the current lane. Additionally, each element in $\mathbf{P}_{y,g}$ is clipped using $\text{clip}({p}^{(j)}_{y,g},p_{y,\min},p_{y,\max})$ to facilitate the EV in adjusting its lateral position to proactively avoid road construction areas.  
\begin{algorithm}[t]
\caption{Sampling Points Generation}\label{alg:alg1}
\begin{algorithmic}[1]
    \State \textbf{Parameters:} 
        \Statex \hspace{0.5cm} $v_d$: Desired longitudinal driving speed;  
        \Statex \hspace{0.5cm} $a_{x,\text{max}}$: Maximum longitudinal acceleration;
        \Statex \hspace{0.5cm} $a_{x,\text{min}}$: Minimum longitudinal acceleration;
        \Statex \hspace{0.5cm} $j_{x,\text{max}}$: Maximum longitudinal jerk;
        \Statex \hspace{0.5cm} $j_{x,\text{min}}$: Minimum longitudinal jerk;
        \Statex \hspace{0.5cm} $\Delta d_x$: Constant adjustment interval;
        \Statex \hspace{0.5cm} $M$: Number of the nearest HVs;
    \State Measure the states of the $M$ nearest HVs: $\left\{ \mathbf{O}^{(i)}_{0} \right\}_{i=1}^{M}$;
    \State Measure the current state of the EV: $\textbf{x}_0$;
    \State Measure the current state of $M$ nearest surrounding HVs: $\left\{ \mathbf{O}^{(i)}_{0} \right\}_{i=1}^{M}$;
    \State Predict the final step position of the $M$ nearest HVs: $\left\{\mathbf{O}^{(i)}_{N} \right\}_{i=1}^{M}$;
    \State \textbf{Update} the target lateral goal vector $ \mathbf{P}_{y,g} $ using \eqref{eq:lateral_goal};
    \State \textbf{Update} the target longitudinal goal vector $ \mathbf{P}_{x,g} $ using \eqref{eq:lon_goal};
    \State Check the safety of each goal point in the goal vector using \eqref{eq:dist};
    \State \textbf{While} \textit{Unsafe} \textbf{do}:
        \Statex \hspace{0.5cm} Decrease the corresponding unsafe longitudinal points 
        \Statex \hspace{0.5cm} with a constant interval $\Delta d_x$;
        \Statex \hspace{0.5cm} Check Safety using \eqref{eq:dist};
    \State \textbf{End While}
    \State Obtain the goal vector $\begin{matrix}
        [\mathbf{P}_{x,g} & \mathbf{P}_{y,g}]
    \end{matrix}^T$. 
\end{algorithmic} 
\end{algorithm}

Drawing inspiration from the double S velocity profile to generate a smooth acceleration profile for stable motion planning~\cite{biagiotti2008trajectory}, we determine the longitudinal goal position ${p}^{(j)}_{x,g},\forall j\in\mathcal{I}_0^{N_c-1}$, based on reachability, task accuracy, and interaction safety between the EV and HVs.

Given the current longitudinal position $p_{x,0}$, velocity $v_{x,0}$, acceleration $a_{x,0}$, desired velocity $v_d$, the acceleration range defined in~\eqref{eq:longitudinal_acc_cons}, and the jerk range specified in~\eqref{eq:longitudinal_jerk_cons} for the EV, we can compute the target longitudinal position using the following criteria:
\begin{itemize}
  \item The final longitudinal acceleration is set to zero.
  \item $j_{x,\max} = - j_{x,\min}$.
\end{itemize}  
As a result, we can obtain analytical solutions for the ideal longitudinal distance based on reachability analysis. For brevity, we consider a task where the initial velocity $v_{x,0}$ is less than the desired velocity $v_d$; and vice versa. Two cases can be distinguished, as follows:
\begin{itemize}
  \item \textbf{Case 1:}  
  The longitudinal acceleration increases linearly from an initial value \(a_0\) to \(a_1\) over the time interval \(t \in [t_0, t_1]\) with the maximum longitudinal jerk \(j_{x, \max}\). Subsequently, it decreases linearly to zero from \(t \in [t_1, t_2]\) with the minimum longitudinal jerk \(j_{x, \min}\) and maintains zero acceleration from \(t \in [t_2, T]\) until the end. 
  \item \textbf{Case 2:} The acceleration increases linearly from an initial value \(a_0\) to the maximum value \(a_{x,\max}\) over the time interval \(t \in [t_0, t_a]\) with the maximum longitudinal jerk \(j_{x, \max}\). It then maintains this maximum constant value from \(t \in [t_a, t_c]\) and decreases linearly to zero from \(t \in [t_c, T]\) with the minimum longitudinal jerk \(j_{x, \min}\).
\end{itemize}
 We can get the corresponding analytical longitudinal distance changes $ \delta p_x$ in Case 1 and Case 2 (see Appendix).
Therefore, we can get the target longitudinal goal position vector for $N_c$ free-end homotopic trajectories, as follows: 
\begin{equation} \vspace{-1mm}
     \mathbf{P}_{x,g} = p_x +  \delta \mathbf{x},
     \label{eq:lon_goal}
 \vspace{-1mm} \end{equation} 
where 
$ \mathbf{P}_{x,g} =  \left\{ p^{(j)}_{x,g} \right\}_{j=1}^{N_c} \in \mathbb{R}^{1\times N_c}$; $p_x$ is the current longitudinal position of the EV; $ \delta \mathbf{x} =  \left\{ \delta p^{(j)}_{x,g} \right\}_{j=1}^{N_c} \in \mathbb{R}^{1\times N_c}$, where the variable \(\delta p^{(j)}_{x,g}\) is initialized to \( \delta p_x \) at the beginning of each receding horizon planning. Additionally, it undergoes dynamic online adjustments to avoid collision-occupied regions with the following safety checking function $h_D$:
\begin{equation} \vspace{-1mm} 
    h_D = \frac{(p^{(j)}_{x,g}- o^{(i)}_{x,N})^2}{a^2_i}+\frac{(p^{(j)}_{y,g} - o^{(i)}_{y,N})^2}{b^2_i}-1, i\in\mathcal{I}_0^{N_c-1},
     \label{eq:dist}
 \vspace{-1mm} \end{equation} 
 where $o^{(i)}_{x,N}$ and $ o^{(i)}_{y,N}$ represent the predicted final longitudinal and lateral position of the $i$-th HVs, respectively; 
 $a_i$ and $b_i$ denote the major and minor axis lengths of a safe ellipse for the interaction of the EV and $i$-th HV, respectively. The positive value of the safety checking function $h_D$ denotes safety; and vice versa. The procedure is detailed in \textbf{Algorithm} \ref{alg:alg1}.

\begin{remark} In Case 2, when the desired velocity \(v_\text{d}\) of the EV significantly exceeds its initial longitudinal speed \(v_{x,0}\), there is a possibility that the EV may not reach the desired velocity \(v_\text{d}\) within its planning horizon \(T\). For instance, the high-speed EV got cut in by a low-speed HV in a cruise task.  
Our approach to determining the goal vector based on double S velocity optimization can facilitate stable driving under this condition.
\end{remark}  
\section{Barrier-Enhanced Parallel Homotopic Trajectory Optimization (BPHTO) with Over-relaxed ADMM}
    \label{subsec:reformulation_PTO} 
    \subsection{Problem Reformulation}
    By incorporating spatiotemporal control barriers, dynamic feasibility, task-oriented motion constraints, 
    we can reformulate the initial NLP \eqref{eq:bp_pto1}--\eqref{eq:bp_pto6} into the following  bi-convex optimization problem:
    \begin{subequations}    
    \label{problem3}
      \begin{align}  
      \displaystyle\operatorname*{min}_{\substack{\{\mathbf{C}_\theta, \mathbf{C}_x, \mathbf{C}_y\}
     \\
    \{\bm{\omega}, \mathbf{d}\}}}~~ 
    \ &\  f(\mathbf{C}_\theta)  + g_x(\mathbf{C}_x) + g_y(\mathbf{C}_y) \label{eq:problem3_pto1}\\
    \operatorname*{s.t.}\quad\quad
    & \ \mathbf{F}_0  \mathbf{C}_{\theta}   \in \mathcal{C}_0,  \mathbf{F}_f    \mathbf{C}_{\theta}  \in \mathcal{C}_f,  \label{eq:problem3_pto2}\\
    &  \mathbf{A}_0 
    \begin{bmatrix} 
   \mathbf{C}_x\\
    \mathbf{C}_y
    \end{bmatrix}  \in \mathcal{D}_0,  \mathbf{A}_f 
    \begin{bmatrix} 
   \mathbf{C}_x\\
    \mathbf{C}_y
    \end{bmatrix} \in \mathcal{D}_f, \label{eq:problem3_pto3}\\     
    & \bm{\omega}  \in \mathcal{C}_{\omega}, \mathbf{d}  \in \mathcal{C}_d, \label{eq:problem3_pto4}\\
    &   \dot{\mathbf{W}}^T_B \mathbf{C}_{x}  - \mathbf{V} \cdot \cos{  \mathbf{W}^T_B\mathbf{C}_{\theta}}  =  0, \label{eq:problem3_pto5}\\     
    &   \dot{\mathbf{W}}^T_B \mathbf{C}_{y}  - \mathbf{V} \cdot \sin{  \mathbf{W}^T_B\mathbf{C}_{\theta}}  =  0, \label{eq:problem3_pto6}\\    
   & \mathbf{V}_{w} \mathbf{C}_{x} - \mathbf{O}_{x}  - \mathbf{L}_x \cdot  \mathbf{d} \cdot  \cos{\bm{\omega}} =  0, \label{eq:problem3_pto7}\\
    &\mathbf{V}_{w}\mathbf{C}_{y}  - \mathbf{O}_{y}  - \mathbf{L}_y \cdot  \mathbf{d} \cdot  \sin{\bm{\omega}} =  0, \label{eq:problem3_pto8}\\ 
    &\mathbf{G} \mathbf{C}_x  -\mathbf{h}_x \leq  0,	\label{eq:problem3_pto9} \\
    &\mathbf{G} \mathbf{C}_y - \mathbf{h}_y  \leq  0.	\label{eq:problem3_pto10}   
\end{align}
\end{subequations} 
   Here, the variable \(\mathbf{C}_\theta \in \mathbb{R}^{(n+1)\times N_c}\) with \([\mathbf{C}_\theta]_{i,j} = c^{(j)}_{\theta, i}\); \(\mathbf{C}_x \in \mathbb{R}^{(n+1)\times N_c}\) with \([\mathbf{C}_x]_{i,j} = c^{(j)}_{x, i}\); \(\mathbf{C}_y \in \mathbb{R}^{(n+1)\times N_c}\) with \([\mathbf{C}_y]_{i,j} = c^{(j)}_{y, i}\). 
    The variable \(\bm{\omega} = [\bm{\omega}_0 \quad \bm{\omega}_1 \quad \dots \quad  \bm{\omega}_{Nc-1}] \in \mathbb{R}^{(N \times M) \times N_c}\), where each \(\bm{\omega}_j \in \mathbb{R}^{N \times M}\) corresponds to a matrix vertically stacking relative angle \(\omega^{(i)}_k\), \(\forall k \in \mathcal{I}_0^{N-1}\), \(i \in \mathcal{I}_0^{M-1}\), considering \(M\) surrounding interactive HVs at \(N\) planning time steps. The variable \(\mathbf{d} = [\mathbf{d}_0\quad \mathbf{d}_1\quad \dots\quad  \mathbf{d}_{Nc-1}] \in \mathbb{R}^{(N \times M) \times N_c}\) has a similar structure, with each \(\mathbf{d}_j \in \mathbb{R}^{N \times M}\) constructed by vertically stacking \(d^{(i)}_k, \forall k \in \mathcal{I}_0^{N-1}\), \(i \in \mathcal{I}_0^{M-1}\).
    
    The sets \(\mathcal{C}_{0}\) and \(\mathcal{D}_{0}\) denote the initial constraints in heading angle, position, and velocity, while the sets \(\mathcal{C}_{f}\) and \(\mathcal{D}_{f}\) denote the corresponding target navigation state sets derived from \eqref{eq:init_pos}--\eqref{eq:final_theta}. \(\mathcal{C}_{\omega}\) and \(\mathcal{C}_d\) in (\ref{eq:problem3_pto4}) represent the value sets of the variables \(d^{(i)}_k\) and \(\omega^{(i)}_k\), respectively. These sets are derived from the constraints specified in (\ref{eq:barrier_cons_polar}) and (\ref{eq:angle_value}), respectively.

   Note that matrix \(\mathbf{V} = [\mathbf{V}_0\quad \mathbf{V}_1\quad \dots\quad \mathbf{V}_{Nc-1}] \in \mathbb{R}^{N\times N_c}\), where each element \(\mathbf{V}\) represents the vertical stacking of \(v^{(j)}_k\) for all \(k \in \mathcal{I}_0^{N-1}\) based on its closed-form solution (\ref{eq:polar_vel}). 
    The constant matrices \(\mathbf{L}_x\) and \(\mathbf{L}_y\) are defined as \(\mathbf{L}_x = [\mathbf{L}_{x,0}\quad \mathbf{L}_{x,1}\quad \dots \quad \mathbf{L}_{x,Nc-1}]\) and \(\mathbf{L}_y = [\mathbf{L}_{y,0}\quad \mathbf{L}_{y,1}\quad \dots \quad \mathbf{L}_{y,Nc-1}]\), both belonging to \(\mathbb{R}^{(N\times M)\times N_c}\). Each element of these matrices is constructed by stacking \(l^{(i)}_x\) and \(l^{(i)}_y\) for \(N\) planning time steps, \(\forall i \in \mathcal{I}_0^{M-1}\).
    
    The predicted longitudinal and lateral positions of \(M\) surrounding HVs in \(N\) planning time steps for \(N_c\) free-end homotopic candidate trajectories are denoted by \(\mathbf{O}_{x} \in \mathbb{R}^{(N \times M) \times N_c}\) and \(\mathbf{O}_{y} \in \mathbb{R}^{(N \times M) \times N_c}\), respectively. Additionally, the constant matrix \(\mathbf{V}_{w} = \mathbf{W}^T_B \otimes \mathbf{I}_{M} \in \mathbb{R}^{(N \times M) \times (n+1)}\) represents the vertical stacking of \(\mathbf{W}^T_B\) for \(M\) surrounding HVs.

    In the objective function (\ref{eq:problem3_pto1}), $ f(\mathbf{C}_\theta)$, $ g(\mathbf{C}_x)$, and $ g(\mathbf{C}_y)$ are in the quadratic form: 
\begin{equation} \vspace{-0.5mm}
  f(\mathbf{C}_\theta) = \frac{1}{2} \mathbf{C}^T_\theta Q_{\theta}\mathbf{C}_\theta,
  \label{eq:f_theta}
 \vspace{-1mm} \end{equation}
\begin{equation} \vspace{-1mm}
  g(\mathbf{C}_x) = \frac{1}{2} \mathbf{C}^T_x Q_{x}\mathbf{C}_x,
  \label{eq:f_x}
 \vspace{-1mm} \end{equation}
\begin{equation} \vspace{-1mm}
  g(\mathbf{C}_y) = \frac{1}{2} \mathbf{C}^T_y Q_{y}\mathbf{C}_y,
  \label{eq:f_y}
 \vspace{-0.5mm} \end{equation} 
where $\mathbf{Q}_{\theta},  \mathbf{Q}_{x}, \mathbf{Q}_{y} \in \mathbb{R}^{(n+1) \times (n+1)}$ are positive definite weighting matrices to penalize high-order derivatives of the generated trajectory~(\ref{eq:discrete_traj}), facilitating smoothness in the optimization process. 

The matrices $\mathbf{F}_0$, $\mathbf{F}_f$, $\mathbf{A}_0$, and $\mathbf{A}_f$ in \eqref{eq:problem3_pto3} and \eqref{eq:problem3_pto4} are constant matrices enforcing initial constraints and facilitating the EV in reaching its local target state when planning: 
\begin{equation} \vspace{-1mm}
 \mathbf{F}_0  =  \mathbf{A}_0  =  \begin{bmatrix}  B_{0,n}(\delta t)\  B_{1,n}(\delta t)\  \dots\  B_{n,n}(\delta t) \\
\dot{B}_{0,n}(\delta t)\ \dot{B}_{1,n}(\delta t)\ \dots\  \dot{B}_{n,n}(\delta t)  \end{bmatrix} \in \rr^{2\times (n+1)}, 
\notag
 \vspace{-1mm} \end{equation} 
 \begin{equation} \vspace{-1mm}\mathbf{F}_f = \begin{bmatrix} B_{0,n}(N\delta t)\ B_{1,n}(N\delta t)\  \dots\  B_{n,n}(N\delta t)  \\
\dot{B}_{0,n}(N\delta t)\ \dot{B}_{1,n}(N\delta t)\  \dots\ \dot{B}_{n,n}(N\delta t) \end{bmatrix} \in \rr^{2\times (n+1)}, 
\notag
  \vspace{-1mm} \end{equation} 
 \begin{equation} \vspace{-1mm} 
 \mathbf{A}_f  =  [ B_{0,n}(\delta t)\   B_{1,n}(\delta t)\ \dots\  B_{n,n}(\delta t) ]  \in \rr^{1\times (n+1)}.
\notag
 \vspace{-1mm} \end{equation}  
 
Also, matrices \(\mathbf{G} \in \mathbb{R}^{6N\times (n+1)}\), \(\mathbf{h}_x \in \mathbb{R}^{6N\times N_c}\), and \(\mathbf{h}_y \in \mathbb{R}^{6N\times N_c}\) are defined as follows: 
\begin{align}
  \mathbf{G} = \begin{bmatrix}\mathbf{W}^T_B& -\mathbf{W}^T_B& \ddot{\mathbf{W}}^T_B& -\ddot{\mathbf{W}}^T_B& \dddot{\mathbf{W}}^T_B& -\dddot{\mathbf{W}}^T_B\end{bmatrix}^T,
  \notag
\end{align}
\begin{equation} \vspace{-1mm}
\mathbf{h}_x = [\mathbf{P}_{x,\max}\ -\mathbf{P}_{x,\min}\ {\mathbf{A}}_{x,\max}\ -{\mathbf{A}}_{x,\min}\ {\mathbf{J}}_{x,\max}\ -{\mathbf{J}}_{x,\min}]^T,
\notag
 \vspace{-1mm} \end{equation} 
\begin{equation} \vspace{-1mm}
\mathbf{h}_y = [\mathbf{P}_{y,\max}\ -\mathbf{P}_{y,\min}\ {\mathbf{A}}_{y,\max}\ -{\mathbf{A}}_{y,\min}\ {\mathbf{J}}_{y,\max}\ -{\mathbf{J}}_{y,\min}]^T,
\notag
 \vspace{-1mm} \end{equation} 
 where matrices $\mathbf{P}_{x,\max}$, $\mathbf{P}_{x,\min}$, $\mathbf{P}_{y,\max}$, $\mathbf{P}_{y,\min}$, $\mathbf{A}_{x,\max}$, $\mathbf{A}_{x,\min}$, $\mathbf{A}_{y,\max}$, $\mathbf{A}_{y,\min}$, $\mathbf{J}_{x,\max}$, $\mathbf{J}_{x,\min}$, $\mathbf{J}_{y,\max}$, $\mathbf{J}_{y,\min} \in \mathbb{R}^{N\times N_c}$ are defined with elements as follows:
\[
\begin{aligned}
    &\mathbf{P}_{x,\max}[k,j] = p_{x, \max}, \quad \mathbf{P}_{x,\min}[k,j] = p_{x, \min}, \\
    &\mathbf{P}_{y,\max}[k,j] = p_{y, \max}, \quad \mathbf{P}_{y,\min}[k,j] = p_{y, \min}, \\
    &\mathbf{A}_{x,\max}[k,j] = a_{x, \max}, \quad \mathbf{A}_{x,\min}[k,j] = a_{x, \min}, \\
    &\mathbf{A}_{y,\max}[k,j] = a_{y, \max}, \quad \mathbf{A}_{y,\min}[k,j] = a_{y, \min}, \\
    &\mathbf{J}_{x,\max}[k,j] = j_{x, \max}, \quad \mathbf{J}_{x,\min}[k,j] = j_{x, \min}, \\
    &\mathbf{J}_{y,\max}[k,j] = j_{y, \max}, \quad \mathbf{J}_{y,\min}[k,j] = j_{y, \min}.
\end{aligned}
\]
Note that the road boundaries for different driving tasks are enforced by the maximum and minimum lateral position $ p_{y, \max}$ and $ p_{y, \min}$ for safety considerations. In the case of emergency driving scenarios, such as encountering road construction, the values of $ p_{y, \min}$ and $ p_{y, \max}$ can be dynamically adjusted online to accommodate this cluttered scenario. 

\begin{remark}
\label{remark:safety}
{As elaborated in~\cite{rastgar2020novel}, the spatiotemporal safety constraint~(\ref{eq:polar_safety2}) exhibits bi-convexity with respect to $[ p_{x,k}\quad p_{y,k}\quad d^{(i)}_k ]^T$ and $[ \cos(\omega^{(i)}_k)\quad \sin(\omega^{(i)}_k) ]^T$.  Hence, the joint constraints \eqref{eq:problem3_pto7} and  \eqref{eq:problem3_pto8} exhibit bi-convexity, which facilitates the decomposition into two subconstraints during optimization. 
One optimizes $[\mathbf{C}_{x}\quad \mathbf{C}_{y}]^T$ and $\mathbf{d}$ over fixed $\bm{\omega}$, while the other optimizes $\bm{\omega}$ over fixed $[ \mathbf{C}_{x} \quad \mathbf{C}_{y}\quad \mathbf{d}]^T$. 
Similarly, in addressing the joint constraints \eqref{eq:problem3_pto5} and \eqref{eq:problem3_pto6}, one optimizes $[\mathbf{C}_{x}\quad \mathbf{C}_{y}]^T$ over a fixed $\mathbf{C}_{\theta}$, while the other optimizes $\mathbf{C}_{\theta}$ over fixed $[\mathbf{C}_{x}\quad \mathbf{C}_{y}]^T$.
The inherent convexity of the objective function \eqref{eq:problem3_pto1}, characterized by individual quadratic terms \eqref{eq:f_theta},  \eqref{eq:f_x}, and \eqref{eq:f_y}, coupled with the separable nature of the constraints, facilitates its decomposition into distinct subproblems. This separability allows us to efficiently leverage ADMM strategies to divide the bi-convex optimization problem \eqref{problem3} into QP subproblems.} \end{remark}

\subsection{Over-Relaxed ADMM}
\label{subsec:ADMM_iteration}
In this study, we solve this bi-convex optimization problem \eqref{problem3} using the over-relaxed ADMM by introducing slack vectors $\mathbf{Z}_x$ and $\mathbf{Z}_y$, and applying an infinite penalty to their negative components. Each iteration includes solving smaller convex subproblems, and the associated dual variables are updated in parallel to accelerate the computation speed.
 To decompose this optimization problem~(\ref{problem3}) using ADMM, we introduce the definition of the indicator function. 
    \begin{definition}\label{def:indicator_func} 
            The indicator function with respect to a set $\mathbb{S}$ is defined as 
     \begin{equation} \vspace{-1mm}
        \mathcal{I}_{\mathbb{S}}(S)   =
    \begin{cases}
    0 & \text{if } S \in \mathbb{S}, \\
    \infty & \text{if } S \notin \mathbb{S}.  
    \end{cases}
    \label{eq:indicator_func}
     \vspace{-1mm} \end{equation} 
	\end{definition}  

  In particular, the associated \emph{augmented Lagrangian} of (\ref{problem3}) can be formulated as follows:  
\begin{align} \vspace{-1mm}
&\mathcal{L}\left( {\mathbf{C}_\theta, \mathbf{C}x, \mathbf{C}_y},   {\bm{\omega}, \mathbf{d}},  {\mathbf{Z}_x, \mathbf{Z}_y},  {\bm{\lambda}_{\theta}, \bm{\lambda}_{x}, \bm{\lambda}_{y}, \bm{\lambda}_{\text{obs},x}, \bm{\lambda}_{\text{obs},y}} 
\right) \nonumber\\
    &= f(\mathbf{C}_\theta) + g_x(\mathbf{C}_x) + g_y(\mathbf{C}_y) + \mathcal{I}_{+}(\mathbf{Z}_x) + \mathcal{I}_{+}(\mathbf{Z}_y) \nonumber\\
    &\quad+ \bm{\lambda}^T_{\theta}  \left( \dot{\mathbf{W}}^T_B \mathbf{C}_{x}  - \mathbf{V} \cdot \cos{ \mathbf{W}^T_B\mathbf{C}_{\theta}} \right) \nonumber\\
    &\quad+ \bm{\lambda}^T_{\theta}  \left( \dot{\mathbf{W}}^T_B \mathbf{C}_{y}  - \mathbf{V} \cdot \sin{ \mathbf{W}^T_B\mathbf{C}_{\theta}} \right) \nonumber\\
    &\quad+ \bm{\lambda}^T_{\text{obs},x}  \left( \mathbf{V}_{w}\mathbf{C}_{x} - \mathbf{O}_{x}  - \mathbf{L}_x \cdot  \mathbf{d} \cdot  \cos{\bm{\omega}} \right) \nonumber\\
    &\quad+ \bm{\lambda}^T_{\text{obs},y}  \left( \mathbf{V}_{w}\mathbf{C}_{y} - \mathbf{O}_{y}  - \mathbf{L}_y \cdot  \mathbf{d} \cdot  \sin{\bm{\omega}} \right) \nonumber\\
    &\quad+ \bm{\lambda}^T_{x}\mathbf{C}_x  + \bm{\lambda}^T_{y}\mathbf{C}_y \nonumber\\
    &\quad+ \frac{\rho_{\theta}}{2}  \left\| \dot{\mathbf{W}}^T_B \mathbf{C}_{x}  - \mathbf{V} \cdot \cos{ \mathbf{W}^T_B\mathbf{C}_{\theta}} \right\|_{2}^{2} \nonumber\\
    &\quad+ \frac{\rho_{\theta}}{2}  \left\| \dot{\mathbf{W}}^T_B \mathbf{C}_{y}  - \mathbf{V} \cdot \sin{ \mathbf{W}^T_B\mathbf{C}_{\theta}} \right\|_{2}^{2} \nonumber\\
    &\quad+ \frac{\rho_{\text{obs},x}}{2}  \left\| \mathbf{V}_{w}\mathbf{C}_{x} - \mathbf{O}_{x}  - \mathbf{L}_x \cdot  \mathbf{d} \cdot  \cos{\bm{\omega}} \right\|_{2}^{2} \nonumber\\
    &\quad+ \frac{\rho_{\text{obs},y}}{2}  \left\| \mathbf{V}_{w}\mathbf{C}_{y} - \mathbf{O}_{y}  - \mathbf{L}_y \cdot  \mathbf{d} \cdot  \sin{\bm{\omega}} \right\|_{2}^{2} \nonumber\\
    &\quad+ \frac{\rho_x}{2} \left\| \mathbf{G} \mathbf{C}_x  -\mathbf{h}_x + \mathbf{Z}_x \right\|_{2}^{2} \nonumber\\
    &\quad+ \frac{\rho_y}{2} \left\| \mathbf{G} \mathbf{C}_y  -\mathbf{h}_y + \mathbf{Z}_y \right\|_{2}^{2}, 
    \label{lang_dual_problem}
\vspace{-1mm} \end{align} 
where $ \bm{\lambda}_{\theta}\in \rr^{N\times N_c}, \bm{\lambda}_{\text{obs},x} \in \rr^{(N\times M)\times N_c}, \bm{\lambda}_{\text{obs},y} \in \rr^{(N\times M)\times N_c}
 $ are dual variables of the equality constraints \eqref{eq:problem3_pto5}--\eqref{eq:problem3_pto8}; $\rho_{\theta}$, $\rho_{\text{obs},x}$, and $\rho_{\text{obs},y}$ are the corresponding $l_2$ penalty parameters. 
The dual variables \( \bm{\lambda}_{x}  \in \rr^{(n+1)\times N_c} \)  and \( \bm{\lambda}_{y} \in \rr^{(n+1)\times N_c} \) are associated with constraints \eqref{eq:problem3_pto9} and \eqref{eq:problem3_pto10}, respectively, contributing to enhanced iteration stability, as discussed in~\cite{taylor2016training}; $\rho_{x}$  and $\rho_{y}$ are the corresponding $l_2$ penalty parameter. To facilitate optimization and get a feasibility solution,
we introduce slack vectors $\mathbf{Z}_x$ and $\mathbf{Z}_y$ to handle inequality constraints \eqref{eq:problem3_pto9} and  \eqref{eq:problem3_pto10}, respectively.  Specifically, the indicator function  $\mathcal{I}_{+}(\mathbf{Z}_x) = 0$ if constraints \eqref{eq:problem3_pto9} is satisfied, and $\mathcal{I}_{+}(\mathbf{Z}_x) = \infty$ otherwise.
Similarly, $\mathcal{I}_{+}(\mathbf{Z}_y)$ is the indicator function of the  constraints in  \eqref{eq:problem3_pto10}.   
The structure of the \emph{augmented Lagrangian} function~\eqref{lang_dual_problem} and the decomposed set constraints~\eqref{eq:problem3_pto2}--\eqref{eq:problem3_pto4} allow us to group the primal variables into five groups $\{\mathbf{C}_\theta \}$, $\{ \mathbf{C}_x, \mathbf{Z}_x\}$, $\{ \mathbf{C}_y, \mathbf{Z}_y\}$, $\{ \bm{\omega} \}$, and $\{\mathbf{d}\}$  for alternating optimization in five subproblems. Finally, the associated dual variables can be updated concurrently. 
\subsubsection{ADMM Iteration for Equality Constraints}
In the ADMM iteration,
it aims at solving the following sub-problem for the primal variable $\mathbf{C}_\theta$:  
\begin{alignat}{2} \vspace{-1mm}
    \mathbf{C}^{\iota+1}_{\theta}  &= \displaystyle\operatorname*{ \min}_{\mathbf{C}_{\theta}}~~
    \mathcal{L}   \Big(  \{\mathbf{C}_\theta\}, \{ \mathbf{C}^{\iota}_x, \mathbf{Z}^{\iota}_x\}, \{ \mathbf{C}^{\iota}_y, \mathbf{Z}^{\iota}_y\},  \{\bm{\omega}^{\iota}\}, \{ \mathbf{d}^{\iota}\}, \nonumber \\
    &\qquad\qquad\qquad \{\bm{\lambda}^{\iota}_{\theta},  \bm{\lambda}^{\iota}_{x}, \bm{\lambda}^{\iota}_{y},  \bm{\lambda}^{\iota}_{\text{obs},x}, \bm{\lambda}^{\iota}_{\text{obs},y} \}   \Big) \label{eq:sub_qp1}\\
    & = \displaystyle\operatorname*{min}_{\mathbf{C}_{\theta}}~~\frac{1}{2} \mathbf{C}^T_\theta Q_{\theta}\mathbf{C}_\theta +  
 \bm{\lambda}_{\theta}^{\iota T}
 \left\| \begin{matrix}  \dot{\mathbf{W}}^T_B \mathbf{C}^{\iota}_{x} - \mathbf{V} \cdot \cos{ \mathbf{W}^T_B\mathbf{C}_{\theta}} \\  \dot{\mathbf{W}}^T_B \mathbf{C}^{\iota}_{y} - \mathbf{V} \cdot \sin{ \mathbf{W}^T_B\mathbf{C}_{\theta}} \end{matrix} \right\|  \notag\\
&\qquad\qquad + \frac{{\rho}_{\theta}}{2}\left\| \begin{matrix}  \dot{\mathbf{W}}^T_B \mathbf{C}^{\iota}_{x} - \mathbf{V} \cdot \cos{ \mathbf{W}^T_B\mathbf{C}_{\theta}} \\  \dot{\mathbf{W}}^T_B \mathbf{C}^{\iota}_{y} - \mathbf{V} \cdot \sin{ \mathbf{W}^T_B\mathbf{C}_{\theta}} \end{matrix} \right\|^2
 \notag\\
   &   \qquad \text{s.t.}\quad
    \mathbf{F}_0  \mathbf{C}_{\theta}  = [\bm{\theta}_0\quad \dot{\bm{\theta}}_0]^T,  \mathbf{F}_f \mathbf{C}_{\theta}  = \mathbf{0}, \nonumber 
\vspace{-1mm} \end{alignat}
where $\iota$ denotes the current iteration number. The target set $ \mathbf{C}_{f} $ is set to a zero vector. This constraint is imposed to ensure that the final step of each trajectory exhibits a zero heading angle and yaw rate, thereby enhancing driving stability. The vectors $\bm{\theta}_0   \in \rr^{1\times N_c}$ and $\dot{\bm{\theta}}_0  \in \rr^{1\times N_c} $ are formed by horizontally stacking the initial heading angle $\theta(0)$ and yaw rate $\dot{\theta}(0)$ for $N_c$ free-end homotopic candidate trajectories.

Leveraging the polar transformation \eqref{eq:polar_nonholonomic_cons}, the subproblem \eqref{eq:sub_qp1} can be converted into the following constrained least squares problem: 
\begin{alignat}{2} \vspace{-1mm}
    \mathbf{C}^{\iota+1}_{\theta}  &= \displaystyle\operatorname*{ \min}_{\mathbf{C}_{\theta}}~~  
  \frac{1}{2} \mathbf{C}^T_\theta Q_{\theta}\mathbf{C}_\theta \notag
  \\
  &\qquad\qquad +  \bm{\lambda}_{\theta}^{\iota T}
\left\|  \mathbf{W}^T_B\mathbf{C}_{\theta} -  
 \arctan \left(\frac{\dot{\mathbf{W}}^T_B  \mathbf{C}^{\iota}_{y} }{ \dot{\mathbf{W}}^T_B \mathbf{C}^{\iota}_{x} }\right) \right\| \notag
  \\
&\qquad\qquad + \frac{{\rho}_{\theta}}{2} \left\| \mathbf{W}^T_B\mathbf{C}_{\theta} -  
 \arctan \left(\frac{ \dot{\mathbf{W}}^T_B \mathbf{C}^{\iota}_{y} }{\dot{\mathbf{W}}^T_B  \mathbf{C}^{\iota}_{x} }\right) \right\| ^2  \nonumber \\
   &  \qquad \text{s.t.}\quad
   \mathbf{F}_0  \mathbf{C}_{\theta}  = [\bm{\theta}_0\quad \dot{\bm{\theta}}_0]^T,  \mathbf{F}_f    \mathbf{C}_{\theta}  = \mathbf{0}.\nonumber 
\vspace{-1mm} \end{alignat} 

As a result, we can obtain the following analytical solutions for the variable $\mathbf{C}_{\theta}$: 
\begin{equation} \vspace{-1mm}
   \mathbf{C}_{\theta} = \mathbf{A}_{\theta}^\dagger \mathbf{b}_{\theta},
   \label{eq:theta_results}
 \vspace{-1mm} \end{equation} 
where 
\( \mathbf{A}_{\theta}^\dagger \) denotes the Moore-Penrose pseudoinverse of \( \mathbf{A}_{\theta} \), 
\[ \mathbf{A}_{\theta} = \begin{bmatrix} \mathbf{Q}_{\theta} + \rho_{\theta}\mathbf{W}_B\mathbf{W}^T_B \\ \mathbf{F}_0 \\ \mathbf{F}_f \end{bmatrix}, \]
\[ \mathbf{b}_{\theta} = \begin{bmatrix} - \mathbf{W}_B \bm{\lambda}_{\theta}  + \rho_{\theta}\mathbf{W}_B\arctan \left(\frac{ \dot{\mathbf{W}}^T_B \mathbf{C}^{\iota}_{y} }{ \dot{\mathbf{W}}^T_B \mathbf{C}^{\iota}_{x} }\right) \\ [\bm{\theta}_0\quad \dot{\bm{\theta}}_0]^T \\ \mathbf{0} \end{bmatrix}. \]  
\subsubsection{Over-Relaxed ADMM Iterations for Inequality Constraints}
We aim at solving the following constrained least squares problem for the variable $\{ \mathbf{C}_x, \mathbf{Z}_x\}$ based on the over-relaxed ADMM iteration:    
\begin{alignat}{2} \vspace{-1mm}
    \mathbf{C}_{x}^{\iota+1} &= \displaystyle\operatorname*{ \min}_{\mathbf{C}_{x}}~~
    \mathcal{L}   \Big(  \{\mathbf{C}^{\iota}_\theta\}, \{\mathbf{C}_x,  \mathbf{Z}^{\iota}_x\}, \{ \mathbf{C}^{\iota}_y, \mathbf{Z}^{\iota}_y\},  \{\bm{\omega}^{\iota}\}, \{ \mathbf{d}^{\iota}\}, \nonumber \\
    &\qquad\qquad\qquad \{\bm{\lambda}^{\iota}_{\theta},  \bm{\lambda}^{\iota}_{x}, \bm{\lambda}^{\iota}_{y},  \bm{\lambda}^{\iota}_{\text{obs},x}, \bm{\lambda}^{\iota}_{\text{obs},y} \}   \Big) \notag\\
    & = \displaystyle\operatorname*{min}_{\mathbf{C}_{x}}~~\frac{1}{2} \mathbf{C}^T_x Q_{x}\mathbf{C}_x + \bm{\lambda}_{x}^{\iota T} \mathbf{C}_{x} 
 \notag\\
&\qquad\qquad  + \bm{\lambda}_{\theta}^{\iota T}  \left( \dot{\mathbf{W}}^T_B \mathbf{C}_{x}  - \mathbf{V} \cdot \cos{ \mathbf{W}^T_B\mathbf{C}^{\iota}_{\theta}} \right)  \notag\\
 &\qquad\qquad  + \bm{\lambda}_{\text{obs},x}^{\iota T}  \left( \mathbf{V}_{w}\mathbf{C}_{x} - \mathbf{O}_{x}  - \mathbf{L}_x \cdot  \mathbf{d}^{\iota} \cdot  \cos{\bm{\omega}^{\iota}} \right)  \notag\\
 &\qquad\qquad  + \frac{\rho_{\theta}}{2}  \left\| \dot{\mathbf{W}}^T_B \mathbf{C}_{x}  - \mathbf{V} \cdot \cos{ \mathbf{W}^T_B\mathbf{C}^{\iota}_{\theta}} \right\|_{2}^{2}  \notag\\
&\qquad\qquad  {+\frac{\rho_{\text{obs},x}}{2} \left\| \mathbf{V}_{w}\mathbf{C}_{x} - \mathbf{O}_{x}  - \mathbf{L}_x \cdot  \mathbf{d}^{\iota} \cdot \cos{\bm{\omega}^{\iota}} \right\|_{2}^{2} } \notag\\
&\qquad\qquad + \frac{\rho_x}{2} \left\| \mathbf{G} \mathbf{C}_x  -\mathbf{h}_x + \mathbf{Z}^{\iota}_x \right\|_{2}^{2} \notag\\
   &   \qquad \text{s.t.}\quad
    \mathbf{A}_0  \mathbf{C}_{x}  = [\mathbf{P}_{x,0}\quad \mathbf{V}_{x,0} ]^T,  \mathbf{A}_f \mathbf{C}_{x}  = \mathbf{P}_{x,g}, \notag
\vspace{-1mm} \end{alignat} 
where the vectors $\mathbf{P}_{x,0} \in \rr^{1\times N_c}$ and $\mathbf{V}_{x,0} \in \rr^{1\times N_c} $ are formed by horizontally stacking the initial longitudinal position $p_{x,0}$ and  longitudinal velocity $v_{x,0}$ for $N_c$ free-end homotopic candidate trajectories. 
As a result, we can get the following analytical solutions for the variable $\mathbf{C}_{x}$: 
\begin{equation} \vspace{-1mm}
   \mathbf{C}^{\iota+1}_{x} = \mathbf{A}_{x}^\dagger \mathbf{b}_{x},
   \label{eq:cx_results}
 \vspace{-1mm} \end{equation} 
where \( \mathbf{A}_{x}^\dagger \) denotes the Moore-Penrose pseudoinverse of \( \mathbf{A}_{x} \),
\[ \mathbf{A}_{x} = \begin{bmatrix} \mathbf{Q}_{x} + \rho_{\theta}\dot{\mathbf{W}}_B\dot{\mathbf{W}}^T_B 
+ \rho_{\text{obs},x}\mathbf{V}^T_w  \mathbf{V}_w
+ \rho_{x}  \mathbf{G}^T  \mathbf{G}
\\ \mathbf{A}_0 \\ \mathbf{A}_f \end{bmatrix}, \]  
\[ \mathbf{b}_{x} = \begin{bmatrix} 
  \mathbf{b}_{x,1}
  \\ [\mathbf{P}_{x,0}\quad \mathbf{V}_{x,0} ]^T \\ \mathbf{P}_{x,g}\end{bmatrix}, \] 
and $\mathbf{b}_{x,1} $ is given by
\begin{alignat}{2} \vspace{-1mm}
 \mathbf{b}_{x,1} = & -\bm{\lambda}_{x}^{\iota}
-\dot{\mathbf{W}}_B\bm{\lambda}_{\theta}^{\iota} 
- \mathbf{V}^T_w\bm{\lambda}_{\text{obs},x}^{\iota } \nonumber \\
    & +\rho_{\theta}\dot{\mathbf{W}}_B \mathbf{V} \cdot \cos{ \mathbf{W}^T_B\mathbf{C}^{\iota}_{\theta}}  
 \notag\\
& +  {\rho_{\text{obs},x}\mathbf{V}^T_w ( \mathbf{O}_{x}  + \mathbf{L}_x \cdot  \mathbf{d}^{\iota} \cdot \cos{\bm{\omega}^{\iota}}} )
 \notag\\
 &  + \frac{\rho_x}{2} \mathbf{G}^T (\mathbf{h}_x-\mathbf{Z}^{\iota}_x ) .\notag 
\end{alignat}  \vspace{-\baselineskip}

The corresponding slack variable vector $\mathbf{Z}_x$, which facilitates optimization and guarantees solving feasibility, is updated as follows:
\begin{alignat}{2} \vspace{-1mm}
\mathbf{Z}^{\iota+1}_x
= &\max  \Big( \mathbf{0}, \displaystyle\operatorname*{ \min}_{\mathbf{Z}_{x}}~~
    \mathcal{L} 
\Big( \{\mathbf{C}^{\iota+1} _\theta\}, \{ \mathbf{Z}_x, \mathbf{C}^{\iota+1}_x\}, \{ \mathbf{C}^{\iota}_y, \mathbf{Z}^{\iota}_y\}, \nonumber \\
& \qquad     \qquad \{\bm{\omega}^{\iota}\}, \{ \mathbf{d}^{\iota}\}, \{\bm{\lambda}^{\iota}_{\theta},  \bm{\lambda}^{\iota}_{x}, \bm{\lambda}^{\iota}_{y},  \bm{\lambda}^{\iota}_{\text{obs},x}, \bm{\lambda}^{\iota}_{\text{obs},y} \}   \Big)\Big),  \notag\\
= &\max  \Big( \mathbf{0},  \mathbf{h}_x - \mathbf{G}\mathbf{C}^{\iota+1}_x  \Big),   
\label{eq:z_x_admm_update}
\vspace{-1mm} \end{alignat}
which leads to the corresponding dual variable { $\bm{\lambda}_x$} update:
\begin{equation} \vspace{-1mm}
 \bm{\lambda}^{\iota+1}_x =  \bm{\lambda}^{\iota}_x + \rho_{x} ( \mathbf{G} \mathbf{C}^{\iota+1}_x  -\mathbf{h}_x + \mathbf{Z}^{\iota+1}_x ).
\label{eq:lambda_x_admm_update}
 \vspace{-1mm} \end{equation}  

To improve the convergence properties of the algorithm, one must also account for past iterates when computing the
next ones. Consider the relaxation of~\eqref{eq:z_x_admm_update} and 
\eqref{eq:lambda_x_admm_update} obtained by replacing
$\mathbf{G}\mathbf{C}^{\iota+1}_x$ in $\mathbf{Z}_x$ the $\bm{\lambda}_x$-updates with $\alpha \mathbf{G}\mathbf{C}^{\iota+1}_x- (1-\alpha) (\mathbf{Z}^{\iota}_x-  \mathbf{h}_x )$. The resulting iterates take the form:   
\begin{subequations} \vspace{-1mm}
       \begin{align} 
       \mathbf{Z}^{\iota+1}_x = &\max  \Big( \mathbf{0},  \mathbf{h}_x - \mathbf{G}\mathbf{C}^{\iota+1}_x  \Big),   \label{eq:z_x_relaxedadmm_update} \\
 \bm{\lambda}^{\iota+1}_x = & \bm{\lambda}^{\iota}_x + \rho_{x} (\alpha_x( \mathbf{G} \mathbf{C}^{\iota+1}_x  -\mathbf{h}_x + \mathbf{Z}^{\iota+1}_x) \notag \\ 
  &+ (1-\alpha_x)(\mathbf{Z}^{\iota+1}_x-\mathbf{Z}^{\iota}_x)). \label{eq:lambda_x_relaxedadmm_update}
       \end{align}  
\end{subequations} 

Similarly, the following iteration results for variable $\mathbf{C}_y$ based on the over-relaxed ADMM iteration are obtained:   
\begin{equation} \vspace{-1mm}
   \mathbf{C}^{\iota+1}_{y} = \mathbf{A}_{y}^\dagger \mathbf{b}_{y},
   \label{eq:cy_results}
 \vspace{-1mm} \end{equation} 
where \( \mathbf{A}_{y}^\dagger \) denotes the Moore-Penrose pseudoinverse of \( \mathbf{A}_{y} \), 
\[ \mathbf{A}_{y} = \begin{bmatrix} \mathbf{Q}_{y} + \rho_{\theta}\dot{\mathbf{W}}_B\dot{\mathbf{W}}^T_B 
+ \rho_{\text{obs},y}\mathbf{V}^T_w  \mathbf{V}_w
+ \rho_{y}  \mathbf{G}^T  \mathbf{G}
\\ \mathbf{A}_0 \\ \mathbf{A}_f \end{bmatrix}, \]  
\[ \mathbf{b}_{y} = \begin{bmatrix} 
  \mathbf{b}_{y,1}
  \\ [\mathbf{P}_{y,0}\quad \mathbf{V}_{y,0} ]^T \\ \mathbf{P}_{y,g}\end{bmatrix}. \] 
Here, the vectors $\mathbf{P}_{y,0} \in \rr^{1\times N_c}$ and $\mathbf{V}_{y,0} \in \rr^{1\times N_c} $ are formed by horizontally stacking the initial lateral position $p_{y,0}$ and lateral velocity $v_{y,0}$ for $N_c$ homotopic candidate trajectories. The term \( \mathbf{b}_{y,1} \) is defined as: 
\begin{alignat}{2} \vspace{-0mm}
 \mathbf{b}_{y,1} = & -\bm{\lambda}_{y}^{\iota}
-\dot{\mathbf{W}}_B\bm{\lambda}_{\theta}^{\iota} 
- \mathbf{V}^T_w\bm{\lambda}_{\text{obs},y}^{\iota } \nonumber \\
    & +\rho_{\theta}\dot{\mathbf{W}}_B \mathbf{V} \cdot \sin{ \mathbf{W}^T_B\mathbf{C}^{\iota}_{\theta}}  
 \notag\\
& +  \rho_{\text{obs},y}\mathbf{V}^T_w ( \mathbf{O}_{y}  + \mathbf{L}_y \cdot  \mathbf{d}^{\iota} \cdot \sin{\bm{\omega}^{\iota}} )
 \notag\\
 &  + \frac{\rho_y}{2} \mathbf{G}^T (\mathbf{h}_y-\mathbf{Z}^{\iota}_y ).\notag 
\end{alignat}\vspace{-\baselineskip} 

The updates for the associated slack variable \( \mathbf{Z}_y \) and dual variable \( \bm{\lambda}_y \) are as follows:
   \begin{subequations} \vspace{-1mm}
       \begin{align}  
       \mathbf{Z}^{\iota+1}_y= &\max  \Big( \mathbf{0},  \mathbf{h}_y - \mathbf{G}\mathbf{C}^{\iota+1}_y \Big),   \label{eq:z_y_relaxedadmm_update} \\
 \bm{\lambda}^{\iota+1}_y = & \bm{\lambda}^{\iota}_y+ \rho_{y} (\alpha_y( \mathbf{G} \mathbf{C}^{\iota+1}_y  -\mathbf{h}_y + \mathbf{Z}^{\iota+1}_y) \notag \\ 
  &+ (1-\alpha_y)(\mathbf{Z}^{\iota+1}_y-\mathbf{Z}^{\iota}_y)). \label{eq:lambda_y_relaxedadmm_update}
       \end{align} 
\end{subequations}  
Note that the relaxation parameters \(\alpha_x\) and \(\alpha_y\) are recommended to be within the range \([1.5, 1.8]\), as detailed in \cite{ghadimi2015optimal,eckstein1994parallel}. In this study, we choose \(1.5\) as the iteration coefficient for both \(\alpha_x\) and \(\alpha_y\). 
  
 \subsubsection{ADMM Iterations for Variables $\bm{\omega}$ and $\mathbf{d}$}  
{Referring to~\cite{adajania2023amswarm}
and considering the constraints in \eqref{eq:barrier_cons_polar} and (\ref{eq:angle_value}), the updated iterates for $\bm{\omega}$ and $\mathbf{d}$  can be expressed as follows:  
\begin{equation} \vspace{-0mm}
       \mathbf{d}^{\iota+1} = \max  \Big( \mathbf{1},   
  \mathbf{1} + (\mathbf{1} -\bm{\alpha}_k)\cdot(\mathbf{d}^{\iota} -\mathbf{1}  )\Big), 
  \label{eq:d_admm_update}
 \vspace{-0mm} \end{equation}  
\begin{equation} \vspace{-0mm}
    \bm{\omega}^{\iota+1} = \arctan\left(\frac{\mathbf{L}_x \cdot (\mathbf{V}_{w}\mathbf{C}^{\iota+1}_{y} - \mathbf{O}_{y})}{\mathbf{L}_y \cdot (\mathbf{V}_{w}\mathbf{C}^{\iota+1}_{x} - \mathbf{O}_{x})}\right),
    \label{eq:omega_admm_update}
 \vspace{-0mm} \end{equation}
{where the augmented barrier coefficient matrix $\bm{\alpha}_k \in \rr^{(N\times M)\times N_c}$ and each element lies within the interval \( \in (0, 1)\).} This parameter represents the aggressiveness of collision avoidance maneuvers and safety recovery as elaborated in Section~\ref{subsec:Safety_Constraints}.
{\begin{remark} The update rule \eqref{eq:d_admm_update} ensures that each element of the matrix \( \mathbf{d} \) remains positive, thus adhering to safety constraints during iterations.
\end{remark}  }
 \subsubsection{Dual Update}
    \begin{equation}  
        \bm{\lambda}^{\iota+1}_{\theta}  = \bm{\lambda}^{\iota}_{\theta} + \rho_{\theta} \left\|  \mathbf{W}^T_B\mathbf{C}^{\iota+1}_{\theta} -  
         \arctan \left(\frac{\dot{\mathbf{W}}^T_B  \mathbf{C}^{\iota+1}_{y} }{ \dot{\mathbf{W}}^T_B \mathbf{C}^{\iota+1}_{x} }\right) \right\| \label{eq:lambda_theta_admm_update},
            \end{equation}      \begin{align}  
        \bm{\lambda}^{\iota+1}_{\text{obs},x}  & = \bm{\lambda}^{\iota}_{\text{obs},x} + \rho_{\text{obs},x} \left( \mathbf{V}_{w}\mathbf{C}^{\iota+1}_{x} - \mathbf{O}_{x} \right. \notag \\
      &\quad\quad\quad\quad\quad \quad\quad \left. - \mathbf{L}_x \cdot  \mathbf{d}^{\iota+1} \cdot  \cos{\bm{\omega}^{\iota+1}} \right) \label{eq:lambda_obsx_admm_update},   \end{align}    \begin{align} 
        \bm{\lambda}^{\iota+1}_{\text{obs},y}  &  = \bm{\lambda}^{\iota}_{\text{obs},y} + \rho_{\text{obs},y} \left( \mathbf{V}_{w}\mathbf{C}^{\iota+1}_{y} - \mathbf{O}_{y} \right. \notag \\
      & \quad\quad\quad\quad \quad\quad\quad \left. - \mathbf{L}_y \cdot  \mathbf{d}^{\iota+1} \cdot {\sin{\bm{\omega}^{\iota+1}} }\right) \label{eq:lambda_obsy_admm_update}. \end{align}
 
Detailed procedure of the BPHTO approach is provided in \textbf{Algorithm} \ref{alg:alg2}. 
\begin{algorithm}[t]
\caption{BPHTO with ADMM}\label{alg:alg2}
\begin{algorithmic}[1]
\State \textbf{Parameters}: $f$: Nonlinear Dubin’s car model~\cite{chen2023interactive};
\Statex \hspace{1.75cm} $\mathbf{Q}_{\theta}$,$\mathbf{Q}_{x}$,$\mathbf{Q}_{y}$: Weights in the cost function;  
\Statex \hspace{1.75cm} $\rho_{\theta}$, $\rho_{x}$, $\rho_{y}$, $\rho_{\text{obs},x}$, $\rho_{\text{obs},y}$: $l_2$ penalty weights;
\Statex \hspace{1.75cm} $N_c$: Number of the candidate trajectories;
\Statex \hspace{1.75cm} $N$: Planning horizon; 	
\Statex \hspace{1.75cm} $K_{\max}$: Number of the maximum iterations;
\Statex \hspace{1.75cm} $m$: Dimension of  B\'ezier curves;
\Statex \hspace{1.75cm} $n$: Order of  B\'ezier curves;
\Statex \hspace{1.75cm} {$\bm{\alpha}_k$: Barrier coefficient matrix;}
\Statex \hspace{1.75cm} $\mathbf{L}_x$, $\mathbf{L}_y$: Stack matrices for safe ellipse: 
\Statex \hspace{1.75cm} $\epsilon^{\text{pri}}$: Stopping criterion value of iteration;
\Statex \hspace{1.75cm} $r_l$: Lateral perception range of the EV;
\Statex \hspace{1.75cm} $M$: Number of the anticipated  HVs;
\State \textbf{Initialize} the states of the $M$ nearest HVs: 
 $\left\{ \mathbf{O}^{(i)}_{0} \right\}_{i=1}^{M}$;
\State \textbf{Initialize} the nominal control point matrices $\mathbf{W}_{P,j}$; 
\State Obtain the local target navigation $  [\mathbf{P}_{x,g} \quad \mathbf{P}_{y,g} ]^T$  from  \textbf{Algorithm} \ref{alg:alg1}; 
\State \textbf{While}  task is not done \textbf{do}: 
\State  \hspace{0.5cm} Measure the current state of the EV: $x_0$;
\State  \hspace{0.5cm} Measure the current state of $M$ nearest surrounding
\Statex \hspace{0.5cm} HVs: $\left\{ \mathbf{O}^{(i)}_{0} \right\}_{i=1}^{M}$ within lateral perception range $r_l$; 
\State  \hspace{0.5cm} Predict the future trajectories of the $M$ nearest 
\Statex \hspace{0.5cm} HVs: $\left\{\mathbf{O}^{(i)}_{k} \right\}_{k=1}^{N},\ i = 1, 2, \cdots, M$;
\State \hspace{0.5cm} \textbf{For} $\iota \gets 0$ \textbf{to} $ K_{\max}$ \textbf{do}:\\
 \hspace{1cm} \textbf{Update} the primal variable $ \mathbf{C}^{\iota+1}_\theta $ \eqref{eq:theta_results}; \\ 
 \hspace{1cm} {\textbf{Update} the primal variable $ \mathbf{C}^{\iota+1}_x $ \eqref{eq:cx_results};} \\
  \hspace{1cm} \textbf{Update} the slack vector $\ \mathbf{Z}^{\iota+1}_x $\eqref{eq:z_x_relaxedadmm_update}; \\
  \hspace{1cm}  {\textbf{Update} the primal variables $\mathbf{C}^{\iota+1}_y$ \eqref{eq:cy_results}}; \\
   \hspace{1cm} \textbf{Update}  the slack vector $\mathbf{Z}^{\iota+1}_y$ \eqref{eq:z_y_relaxedadmm_update}; \\ 
  \hspace{1cm}  \textbf{Update} the primal variables $ \mathbf{d}^{\iota+1} $ \eqref{eq:d_admm_update}, $ \bm{\omega}^{\iota+1}$  \eqref{eq:omega_admm_update};\\
   \hspace{1cm} \textbf{Update} the dual variables $\bm{\lambda}^{\iota+1}_{\theta}$ \eqref{eq:lambda_theta_admm_update}, $\bm{\lambda}^{\iota+1}_x$ \eqref{eq:lambda_x_relaxedadmm_update},
   \Statex  \hspace{1cm} $\bm{\lambda}^{\iota+1}_y$ \eqref{eq:lambda_y_relaxedadmm_update}, $\bm{\lambda}^{\iota+1}_{\text{obs},x}$ \eqref{eq:lambda_obsx_admm_update}, 
   and $\bm{\lambda}^{\iota+1}_{\text{obs},y}$ \eqref{eq:lambda_obsy_admm_update}; 
\State \hspace{1cm} \textbf{Break} if the primal residual less than $\epsilon^{\text{pri}}$ ; 
\State \hspace{0.5cm} \textbf{End For} 
\State \hspace{0.5cm} Get optimized control point matrices $\mathbf{W}_{P,j}$ and
\Statex \hspace{0.5cm} 
 trajectories $\left\{ \mathbf{C}^{(j)}_k,\dot{\mathbf{C}}^{(j)}_k ,\ddot{\mathbf{C}}^{(j)}_k  \right\}_{k=0}^{N-1}, j\in\mathcal{I}_0^{N_c-1}$; 
\State \hspace{0.5cm} \textbf{Evaluate} each candidate trajectory sequence \eqref{eq:optimal_cost_func};
\State \hspace{0.5cm} \textbf{Select} the optimal trajectory sequence
\Statex \hspace{0.5cm} $\left\{ \mathbf{C}^{(j*)}_k ,\dot{\mathbf{C}}^{(j*)}_k ,\ddot{\mathbf{C}}^{(j*)}_k  \right\}_{k=0}^{N-1}$ \eqref{eq:optimal_behavior};
\State \hspace{0.5cm} Send the first step of the optimal trajectory to the EV.
\State \textbf{End While}
\end{algorithmic}
\label{alg2}
\end{algorithm}  
\subsection{Candidate Trajectories Evaluation and Selection}
\label{subsec:decision_making}
In this study, the evaluation algorithm is designed to evaluate all free-end homotopic candidate trajectories obtained in Section \ref{subsec:reformulation_PTO}, then the optimal trajectory $ \xi^*$ aligning with the optimal maneuver $\tau^*$ selected for the EV to execute as follows: 
 \begin{equation} \vspace{-1mm}\label{eq:optimal_cost_func}
	\begin{aligned}
    s(\xi_j, \tau_j)= \textbf{w}^T \textbf{f}(\xi_j, \tau_j),
	\end{aligned}
 \vspace{-1mm} \end{equation}
 \begin{equation}\label{eq:optimal_behavior}
	\begin{aligned}
    \{ \xi^*, \tau^*\}	= \displaystyle\operatorname*{ \min}_{\substack{\xi_j \in \Xi,\tau_j \in \mathcal{T}}}~~ \displaystyle s(\xi_j, \tau_j),
	\end{aligned}
\end{equation}
where $s(\xi, \tau)$ is the overall evaluation cost function. The weight vector $\textbf{w} = [w_g\quad w_l\quad w_s\quad w_c\quad w_m]^T$ indicates the relative significance of each sub-cost for a trajectory. Concurrently, the vector $\textbf{f}(\xi_j, \tau_j )$ denotes a vector of sub-costs that captures various aspects of the $j$-th trajectory's performance as follows:
\begin{equation} \vspace{-0mm}\notag 
 \textbf{f}(\xi_j, \tau_j) = [F_g\quad  F_l\quad F_s\quad   F_c\quad F_m]^T,
 \vspace{-0mm} \end{equation}
where $F_g$, $F_l$, $F_s$, $F_c$, and $F_m$ represent the goal-tracking, lateral deviation, safety, comfort, and consistency costs, respectively. For a cruise driving task, $F_g$ measures the gap between the actual velocity of the EV and the target cruise velocity $v_g$. The lateral deviation cost $F_l$ is computed from the deviation of the target lane of each candidate trajectory and the generated trajectory.
The safety cost $F_s$ is measured by the primal residual of safety in updating dual variables $\lambda_{\text{obs},x}$ \eqref{eq:lambda_obsx_admm_update} and $\lambda_{\text{obs},y}$  \eqref{eq:lambda_obsy_admm_update}. The comfort cost $F_c$ is obtained from the average jerk value of each candidate trajectory. Besides, we leverage a decaying strategy with respect to planning steps in a horizon for $F_g$, $F_l$, and $F_C$ based on our previous work~\cite{zheng2023real}. To enhance driving consistency, the latest selected trajectory is given precedence. Therefore, the consistency cost is measured by the target driving lane changing between two consecutive decision-making instants, as detailed in~\cite{zheng2023real}.

 \section{Experimental Results } 
	\label{sec:sim}  
 In this section, we validate the effectiveness of the proposed BPHTO approach under various cluttered static and dynamic scenarios with uncertain HVs with synthetic IDM and real-world datasets.   
 Our experiments were implemented in C++ and Robot Operating System 2 on an Ubuntu 22.04 system with an AMD Ryzen 5 5600G CPU with six cores @3.90 GHz and 16 GB RAM. The frequency of the processor is at a base clock speed of 2.28 GHz, with a maximum boost frequency of 3.20 GHz and a minimum frequency of 1.20 GHz.  

   \subsection{  Experimental Setup}
   \subsubsection{Dataset}
The efficacy of the proposed BPHTO method in safety-critical driving scenarios is compared with other state-of-the-art baselines under various tasks. We leverage both synthetic IDM and real-world datasets obtained from the Next Generation Simulation (NGSIM) project\footnote{\url{https://data.transportation.gov/Automobiles/Next-Generation-Simulation-NGSIM-Vehicle-Trajector/8ect-6jqj}} in our experiments. The NGSIM dataset was collected from the I-80 freeway in the San Francisco Bay area.
 
\begin{itemize}
    \item IDM dataset: We adopt the IDM simulation model from \cite{adajania2022multi,albeaik2022limitations}, where HVs drive on a one-direction road. The road width is set to 3.75\,\text{m}.
    To ensure the HVs do not engage in stronger acceleration and braking than the maximum deceleration capacity of the EV, the maximum and minimum longitudinal acceleration of HVs are set to $3\,\text{m/s}^2$ and $-4\,\text{m/s}^2$, respectively. The initial and desired longitudinal velocity ranges of HVs are set to [7\,\text{m/s}, 22\,\text{m/s}] and [7\,\text{m/s}, 28.5\,\text{m/s}] with a random setting for cruise and racing tasks, respectively. The number of HVs is set to 18, positioned within the longitudinal range from -50\,\text{m}  to 130\,\text{m} relative to the longitudinal position of the EV. Additionally, we initialize their states at the starting point of a fixed, safe lane with zero acceleration and steering angle. 

   \item The NGSIM dataset\footnote{\url{https://shorturl.at/aLX03}}  consists of 46 HVs. The HVs drive on a six-lane, one-direction road, where the road width is set to 4\,\text{m} for our experiments.
    Collected from the I-80 freeway in the San Francisco Bay area, the dataset exhibits multi-modal driving behaviors, including lane changing and instances of urgent acceleration and deceleration. The data was captured at a timestep of $0.08 \,\text{s}$, ensuring a high temporal resolution for the experiments. 
\end{itemize}
  \begin{table}[tp]
    \centering
    \scriptsize
    \caption{Parameters in The Driving Experiments} 
    \label{table:Parameter_Settings}
    \begin{tabular}{c c c}
    \hline
    Description         & Parameter with value   \\ \hline 
    Front axle distance to center of mass &   $l_f = 1.06 \,\text{m}$\\
    Rear axle distance to center of mass &  $l_r =  1.85 \,\text{m}$\\
    Safety checking parameters &  $a_i= 5.5\,\text{m}$,  $b_i=  4 \,\text{m}$ \\
    Lane width using IDM datasets &  $w_l =  3.75 \,\text{m}$\\
    Lane width using NGSIM datasets &  $w_l =  4 \,\text{m}$\\
    Lateral perception range & $r_p = 8 \,\text{m}$\\
    Anticipated number of nearest HVs & $M=5$\\
    Order and dimension of  B\'ezier curves  &  $n=10$,  $m=3$ \\
    Longitudinal position range  &  $p_{x} \in [-500\,\text{m}, 1000\,\text{m}] $\\
    Longitudinal velocity range  &  $v_{lon } \in [0\,\text{m/s}, 24\,\text{m/s}] $\\
    Longitudinal acceleration range  &  $a_{x} \in [-4\,\text{m/s}^2, 3\,\text{m/s}^2] $\\
    Lateral acceleration range  &  $a_{y} \in [-2\,\text{m/s}^2, 2\,\text{m/s}^2] $\\
    Longitudinal jerk range  &  $j_{x} \in [-2\,\text{m/s}^3, 2\,\text{m/s}^3] $\\
    Lateral jerk range  &  $j_{y} \in [-1.5\,\text{m/s}^3, 1.5\,\text{m/s}^3] $\\
    $l_2$ penalty parameter &  $\rho_{\theta} = \rho_{x} = \rho_{y} = \rho_{\text{obs},x} =  \rho_{\text{obs},y} =5$ \\
    Trajectory evaluation vector &  $\textbf{w} = [200\quad 20\quad 40\quad 20\quad 20]^T$\\ 
    \hline 
    \end{tabular}
    \end{table} 
    \subsubsection{Parameters and Baselines}
    \label{sec:setup}
    \vspace{0mm}
   For each detected HV, we just employ a simple constant velocity prediction model to predict its motion in \textbf{Algorithm} \ref{alg:alg1} and \textbf{Algorithm} \ref{alg:alg2}. Although more state-of-the-art motion prediction models could be implemented for surrounding HVs, our constant prediction model aims to showcase the real-time adaptability and robustness of the proposed BPHTO approach to various uncertain driving scenarios. Each element in the smoothness weighting matrices $\mathbf{Q}_{\theta}$, $\mathbf{Q}_{x}$ and $\mathbf{Q}_{y}$ is set to 200, 100, and 100 respectively; $K_{\max} = 150$; $\epsilon^{\text{pri}} = 1$. All initial value in dual variable vectors $ \bm{\lambda}_{\theta}, \bm{\lambda}_{\text{obs},x}, \bm{\lambda}_{\text{obs},y}, \bm{\lambda}_{x}, \bm{\lambda}_{y}$ are set to zero. The initial barrier coefficient parameter $\alpha_0$ is set to 0.2, which linearly increases to 1 along the planning horizon $N$, resulting in $\alpha_{N-1} =1$.
     The desired lateral position range regarding road boundaries is set to $p_{y} \in [-8\,\text{m}\quad 8\,\text{m}] $.
    Each element in the $\mathbf{L}_x$ and $\mathbf{L}_y$ matrices is configured to be $6\,\text{m}$ and $5.5\,\text{m}$ for constructing the BFs, respectively.  
    The maneuver adjustment vector $\delta \mathbf{y} = [-6\quad -3\quad 0\quad 3\quad 6]$.
    
     When interacting with HVs with the IDM dataset, the target longitudinal and lateral jerk range of the EV are set to $[-0.9\,\text{m}/\text{s}^3, 0.9\,\text{m}/\text{s}^3]$ and $[-0.6\,\text{m}/\text{s}^3,0.6\,\text{m}/\text{s}^3]$ for driving stability consideration. For NGSIM datasets with highly maneuverable and uncertain HVs and static scenarios with higher relative speeds, the longitudinal jerk ranges are configured with larger values, specifically $[-1.5\,\text{m}/\text{s}^3,1.5\,\text{m}/\text{s}^3]$ and $[-1.0\,\text{m}/\text{s}^3, 1.0\,\text{m}/\text{s}^3]$, respectively. However, for stable merging, the longitudinal and lateral jerk sets are set to $[-0.9\,\text{m}/\text{s}^3,0.9\,\text{m}/\text{s}^3]$ and $[-0.6\,\text{m}/\text{s}^3, 0.6\,\text{m}/\text{s}^3]$, respectively.  
    The planning horizon in IDM and NGSIM datasets is set to $N=50$ and $N=60$, respectively. 
    The initial longitudinal velocity is set to $15\,\text{m/s}$ with a zero heading angle for all scenarios. 
    The communication and control frequency for IDM and NGSIM datasets are set to 0.1\,\text{Hz} and 0.08\,\text{Hz}, respectively.
    Other parameters of the experiments are presented in Table~\ref{table:Parameter_Settings}.  
      
        \begin{table}[t]
            \centering
            \scriptsize
            \caption{Performance Comparison Among Different Algorithms in Dense and Cluttered Static Scenario}
            \label{tab:cruise_static_results}    
            \begin{tabular}[c]{|c |  *{1}{c} | *{3}{c}  |}
                \hline
                \multirow{2}{*}{\textbf{Algorithm}} &  
                \multicolumn{1}{c|}{\textbf{ Accuracy}} &
                \multicolumn{3}{c|}{\textbf{ Stability}}  \\
                  &  \mobility& \cruiseconsistency  \\
                \hline
                        %
                \multirow{1}{*}{Batch-MPC }
               & 14.78    &1.62   & 27.06 & 18  \\
                \hline           
                %
                \multirow{1}{*}{BTO}
                & 15.25   & 0.46  & 1.40 & --  \\     
                 \hline
                \multirow{1}{*}{\textbf{BPHTO}}
                & \textbf{15.02}  & \textbf{0.33} & \textbf{1.40} & \textbf{0.57} \\
                \hline
            \end{tabular}    \vspace{-0mm}
        \end{table}
         	\begin{figure}[tp]
		\centering
    	  \includegraphics[scale=0.35]{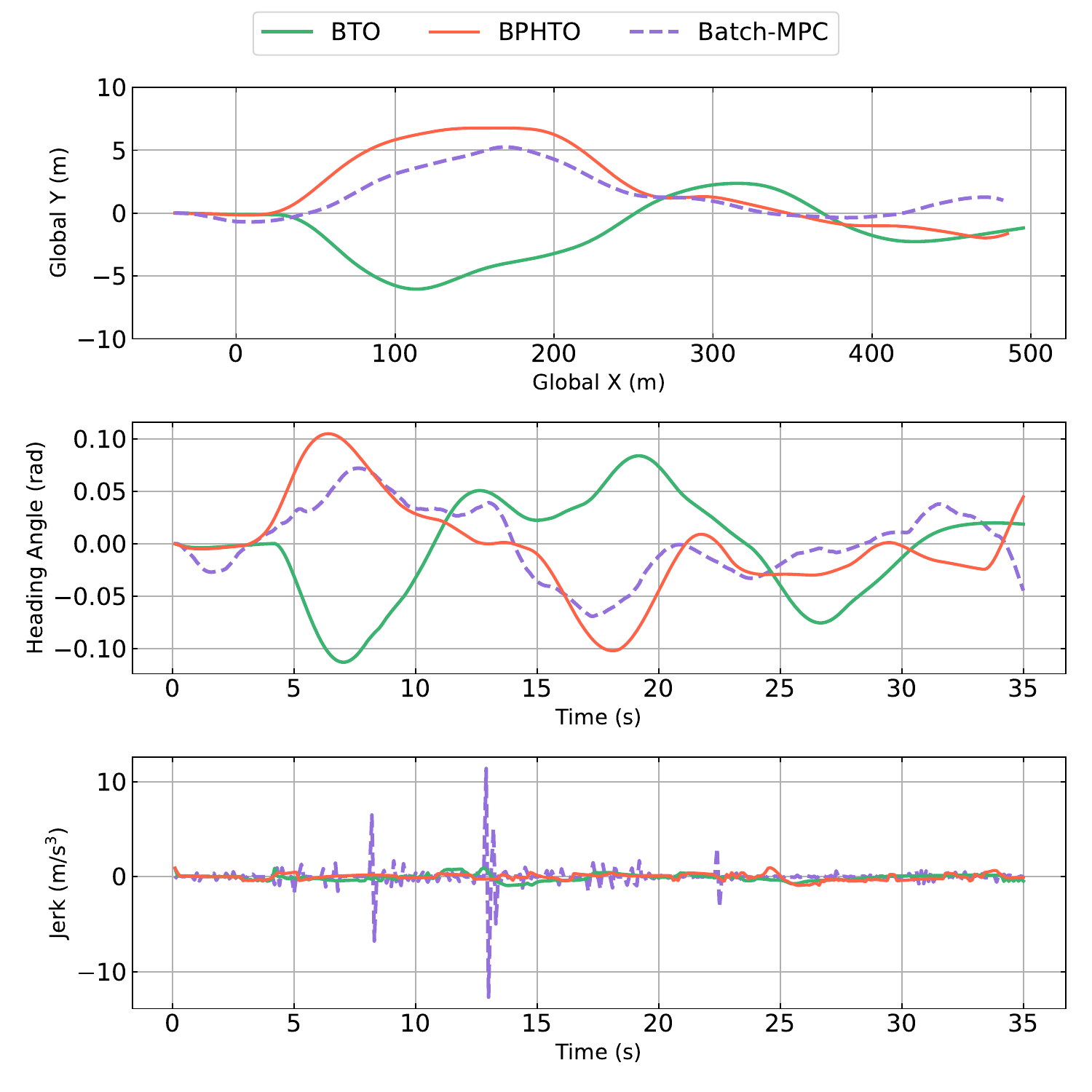}	\vspace{-2mm}
		\caption{{Comparison of position, heading angle, and longitudinal jerk profiles when executing a cruise task using IDM dataset for surrounding HVs’ motion. The evolution of the heading angle profiles reveals how the EV adjusts its orientation to navigate through dense and dynamic traffic. }}\vspace{-2mm}
	\label{fig:NGSIM_cruise_performance}
	\end{figure}
 
\begin{table*}[t]
    \centering
    \scriptsize
    \caption{Performance Comparison Among Different Algorithms in IDM and NGSIM Datasets }
    \label{tab:cruise_dynamic_results}
    \begin{tabular}{|c | *{1}{c} | *{3}{c}|  *{1}{c} | *{3}{c} |}
        \hline
        \multirow{2}{*}{\textbf{Algorithm}} & \multicolumn{4}{c|}{\textbf{IDM dataset}} & \multicolumn{4}{c|}{\textbf{NGSIM dataset}} \\
        \cline{2-9}
         & \textbf{  Accuracy} 
        &   \multicolumn{3}{c|}{\textbf{  Stability}}  & \textbf{  Accuracy} &   \multicolumn{3}{c|}{\textbf{  Stability}}  \\
       & \mobility & \cruiseconsistency  & \mobility & \cruiseconsistency \\
        \hline
        %
        Batch-MPC  & 14.94 & 0.38 & 12.68 & 22.57 & 15.060 &1.58 & 35.52  & 35.43 \\
        \hline
        BTO  & 15.31 & 0.27 & \textbf{0.94} & -- & 15.408  & 0.40 & 1.496 & -- \\
        \hline
        \textbf{BPHTO}  & \textbf{15.02}&  \textbf{0.25}  & 0.95 & \textbf{0.57} & \textbf{14.997} & \textbf{0.26} & \textbf{1.46} & \textbf{0.86} \\
        \hline
    \end{tabular}
    \vspace{-2mm}
\end{table*}
 
  To validate the effectiveness of the proposed BPHTO scheme, we compare it against two baselines. The first baseline is an ablated version of BPHTO, denoted as BTO, which includes only one trajectory without free-end homotopic trajectories. The second baseline is a multi-modal MPC algorithm: Batch-MPC~\cite{adajania2022multi} tailored for highway scenarios based on Bergman alternating minimization. Open-source code\footnote{\url{https://github.com/vivek-uka/Batch-Opt-Highway-Driving}} is leveraged to configure parallel trajectories and optimize parameters for optimal performance. 
   
     \subsection{Comparative Results}
       In this section, we compare the performance of our algorithm with other baselines under static and dynamic dense traffic in cruise tasks. The target longitudinal velocity is set to $v_\text{d} = 15 \,\text{m}/\text{s}$. The assessment criteria encompass several key indicators, including the average cruise speed, mean absolute longitudinal jerk value, and the rate of frequent lane changes between two consecutive decision-making instants.
       
	\subsubsection{Navigation in Static Environments} 
      This EV is required to navigate through a dense obstacle scenario with relatively high speeds to static obstacles.  Each section within the longitudinal range from -50 to 130\,\text{m} relative to the longitudinal position of the EV is configured with ten vehicle-shaped obstacles.  
      The initial position of the EV is set to $[-20\,\text{m}\quad 0\,\text{m} ]$.
      The target lateral for each trajectory of Batch-MPC is set to $[-7.5\quad -3.75\quad 0\quad 3.75\quad 7.5]$, with each optimized trajectory corresponding to the centerline of a driving lane.  The simulation time is set to 30\,\text{s}. 
        
        Table~\ref{tab:cruise_static_results} presents the static performance comparison results of three algorithms in static environments. The results reveal that BPHTO achieves more stable driving behaviors with a lower jerk value and better driving consistency, quantified by the rate of frequent lane changes $\mathcal{P}_{d} = 0.57\,\%$. In contrast, Batch-MPC exhibits a frequent lane changes rate $\mathcal{P}_{d} =18\,\%$, and the largest jerk value is up to around 27$\,\text{m/s}^3$, showing aggressive driving behaviors. Besides, BPHTO achieves the best cruise accuracy among the three algorithms with respect to the desired cruise speed of $15\,\text{m/s}$. Compared with BTO, BPHTO demonstrates improved cruise accuracy, suggesting that employing multiple free-end homotopic trajectories for decision-making enhances task accuracy.

   \subsubsection{ Navigation in Dynamic Dense Traffic}  
      \label{sussubsec:dynamic_cruise}
  We further compare the performance of three different algorithms using synthetic IDM and real-world NGSIM datasets. 
 The simulation employs a time step of 350, resulting in simulation durations of 35\,\text{s} and 28\,\text{s} with the IDM and NGSIM datasets, respectively. The target lateral for Batch-MPC is set to the centerline position of each driving lane. The initial position of the EV is set to $[-40\,\text{m}\quad 0\,\text{m}]$ and $[70\,\text{m}\quad6 \,\text{m}]$ with IDM and NGSIM datasets, respectively.
        
    Table~\ref{tab:cruise_dynamic_results} shows that the average cruise speed of the EV based on BPHTO is closest to the desired cruise speed $15 \,\text{m/s}$ among three algorithms, indicating superior cruise accuracy. Notably, Batch-MPC exhibits a larger maximum longitudinal jerk value ($\mathcal{J}_{mean} = 12\,\text{m/s}^3$) and a higher percentage of the rate of frequent lane changes ($\mathcal{P}_d=22.57\,\%$), indicating that the EV frequently changes its target goal lane, resulting in inferior cruise stability and driving consistency. This finding is supported by the evolution of the longitudinal jerk and acceleration profile depicted in Fig.~\ref{fig:NGSIM_cruise_performance}.
     
   Compared with BTO algorithm, BPHTO achieves better cruise accuracy in both static and dynamic cruise cases using IDM and NGSIM datasets. These observations demonstrate that BPHTO with multiple free-end homotopic trajectories efficiently explores different driving lanes and significantly improves cruise performance.

   Overall, these results showcase the effectiveness of BPHTO in ensuring both high task performance and safety amid challenging dense traffic flow, utilizing IDM and real-world traffic datasets.  
   
    \subsection{Safety Evaluation}
    \subsubsection{Reacting to Road Construction}
    To further highlight the capabilities of our proposed BPHTO to adapt to varying driving environments, we designed a road construction scenario within a cruise task under dense traffic. HVs are controlled in the same manner as described in Section~\ref{sussubsec:dynamic_cruise} using the IDM dataset. Navigating through such cluttered environments presents a substantial challenge for autonomous vehicles~\cite{chen2022milestones, sha2023languagempc}. Due to the inability of other baselines to handle this situation, we do not present their results here.
    The initial position vector of the EV is set to $[0\,\text{m}\quad 5\,\text{m}]^T$, and its front lane is under construction starting at a longitudinal position of 150\,\text{m}, with lateral positions ranging from 1.875\,\text{m} to 9.375\,\text{m}.
    \begin{figure}[tp]
        \centering
        \includegraphics[scale=0.31]{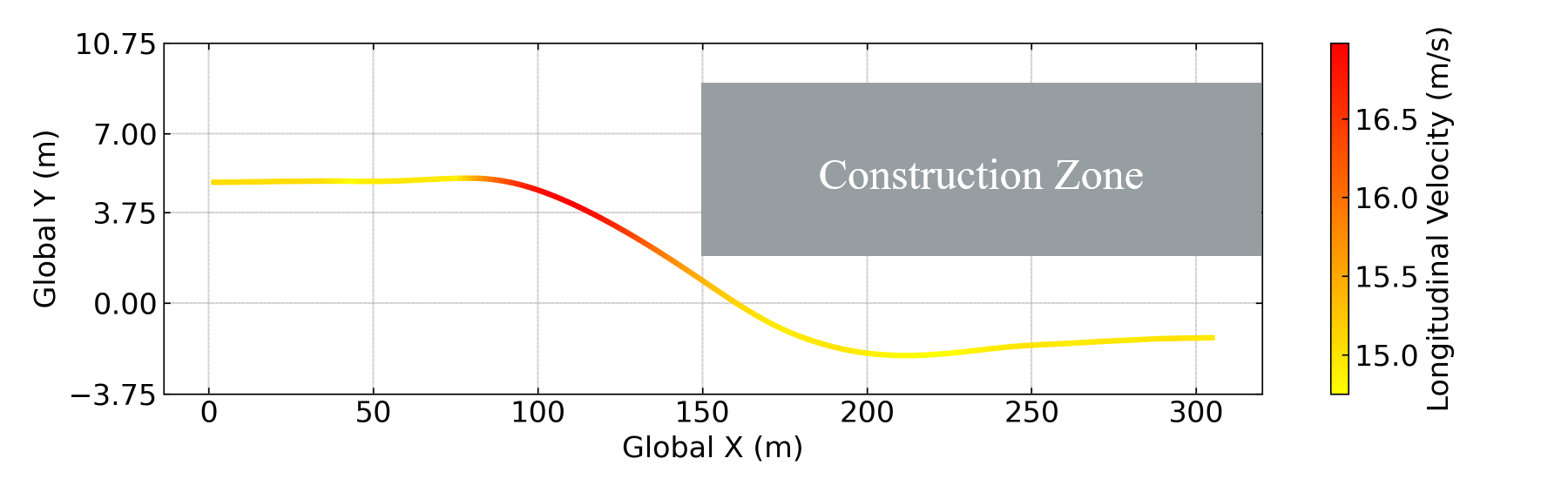} \vspace{-2mm}
        \caption{Evolution of the trajectory in the road construction scenario when executing a cruise task using IDM dataset for surrounding HVs' motion. The motion trajectory of the EV over a simulation duration of $20\,\text{s}$ is colored according to its longitudinal speed profile (yellow-red color).} \vspace{-2mm}
        \label{fig:Cruise_road_construction}
    \end{figure}
  
    The motion trajectory of the EV in Fig.~\ref{fig:Cruise_road_construction} demonstrates its ability to maneuver and avoid the road construction area ahead. Notably, a change in trajectory orientation and an increase in longitudinal velocity around longitudinal position 75$\,\text{m}$ indicate the EV proactively adjusts its orientation and speed to move swiftly into the construction-free lane below. 
    Additionally, the cruise speed consistently maintains proximity to the targeted speed of 15$\,\text{m/s}$. This outcome further underscores the EV's capacity to proactively adjust its driving lane while keeping a stable cruise speed. 
 
\begin{figure}[tp]
		\centering
    	  \includegraphics[scale=0.35]{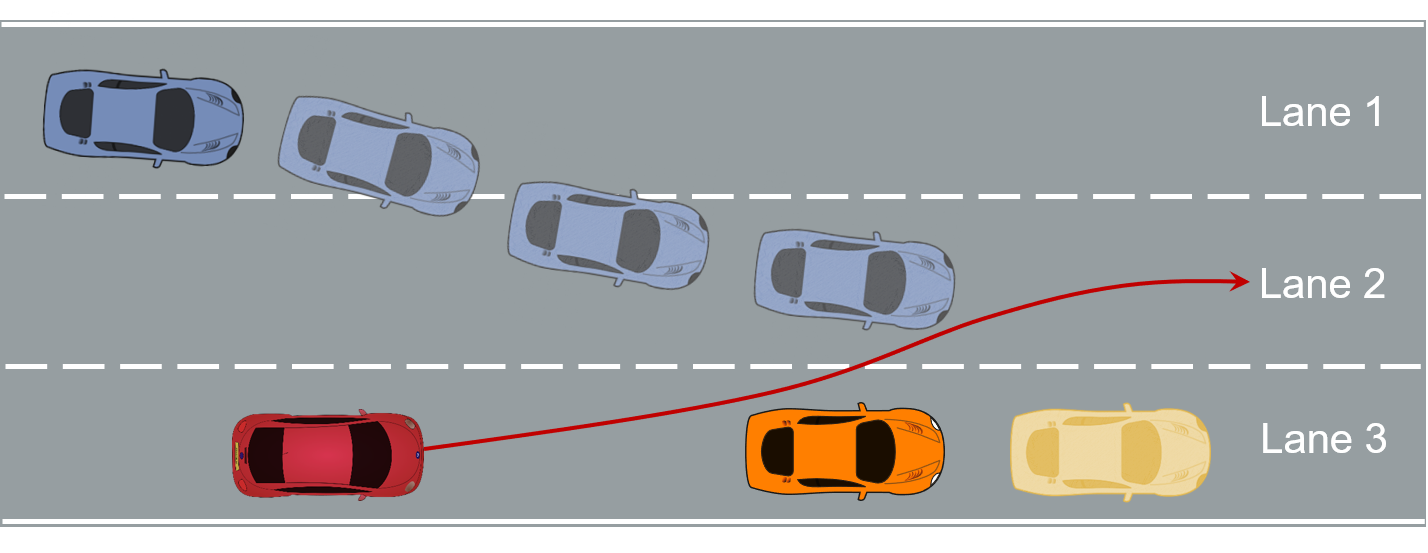}	
		\caption{A lane-merging conflict scenario. The red EV in lane 3 is in the process of executing a lane change to lane 2, while an imperceptible blue HV2 in lane 1 is initiating a lane change to lane 2 from a different direction, potentially leading to a hazardous situation. }
		\label{fig:Cutin_diag}
	\end{figure} 
 
	\subsubsection{Safety Recovery}
        \label{sussubsec:IDM_cruise} 
	\vspace{0mm}
	\label{overtaking_subsec}  
    This subsection assesses the performance of BPHTO in terms of driving stability and safety recovery during a challenging lane-merging conflict
    scenario under a cruise task. The motion of HVs is derived from real-world data in the NGSIM dataset with larger control limits than the EV.  
    The simulation time is set to 19.2\,\text{s}, spanning 240 steps and divided into two distinct phases. In Phase 1, the autonomous red EV endeavors to change lanes and merge with the upper lane 2, where the nearest detected HV is the orange HV1. Simultaneously, an imperceptible blue HV2 in lane 1 executes a lane change to lane 2, as depicted in Fig.~\ref{fig:Cutin_diag}. Phase 2 focuses on the EV's response to the cut-in by HV2, requiring a stable speed adjustment to recover and maintain a safe following distance exceeding 20\,\text{m}. 
	\begin{figure}[tp]
		\centering
    	  \includegraphics[scale=0.365]{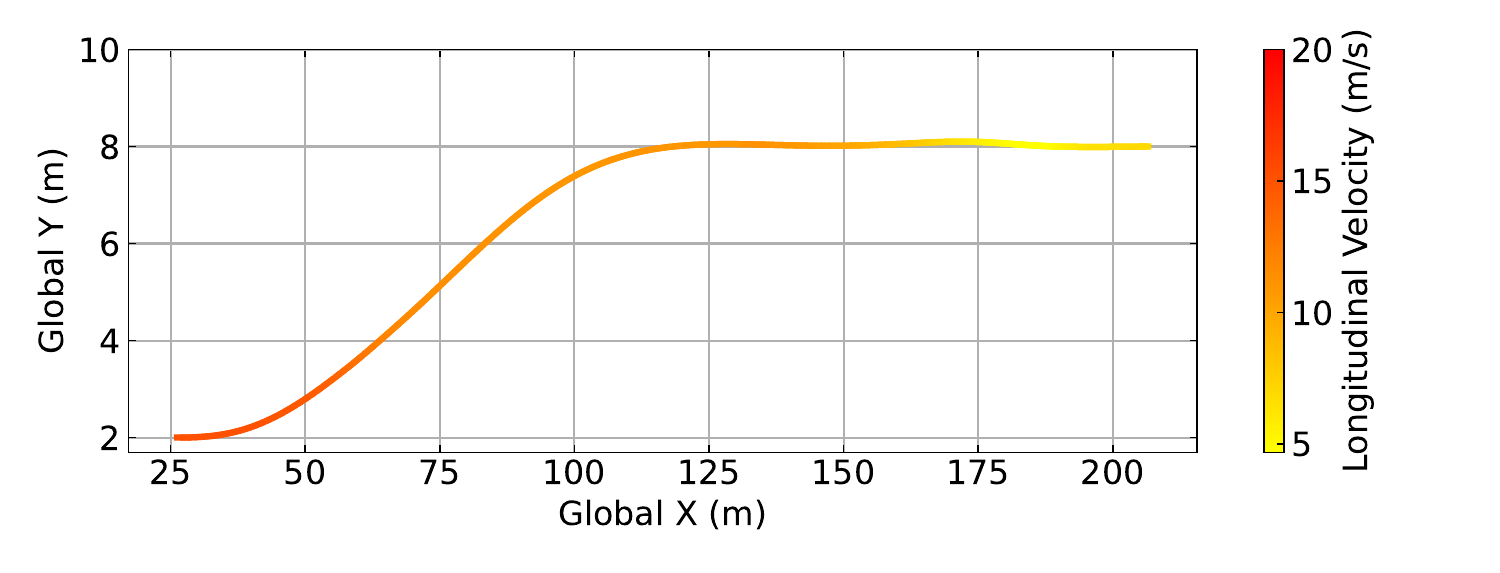} \vspace{-2mm}
		\caption{Evolution of trajectory when executing a lane merge task under cruise scenario using NGSIM dataset for surrounding HVs' motion. It displays the trajectory of the EV, colored according to its longitudinal speed profile (yellow-red color). }\vspace{-0mm}
		\label{fig:NGSIM_cruise_cutin_traj}
	\end{figure} 
        \begin{figure}[tp]
		\centering
    	  \includegraphics[scale=0.35]{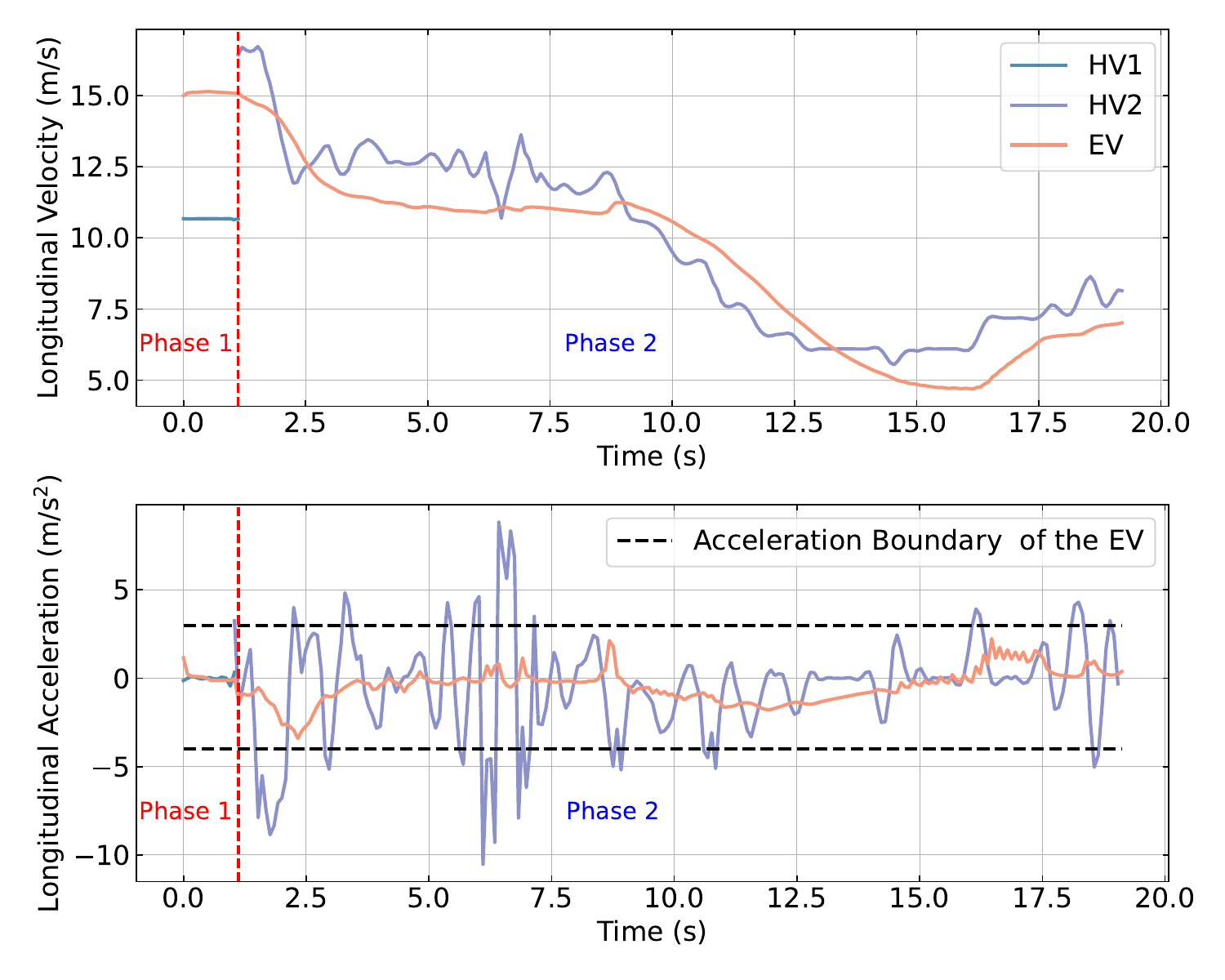}	\vspace{-2mm}
		\caption{Evolution of velocity and acceleration profiles of the EV and its nearest front HVs when executing a lane merge task using NGSIM dataset for surrounding HVs' motion. The similarities in the velocity profile indicate how the EV attempts to adjust its velocity to keep a desired following distance with its front HV.}\vspace{-2mm}
		\label{fig:NGSIM_cruise_cutin_vel_acc}
	\end{figure}

   \begin{figure}[tp]
		\centering
    	  \includegraphics[scale=0.355]{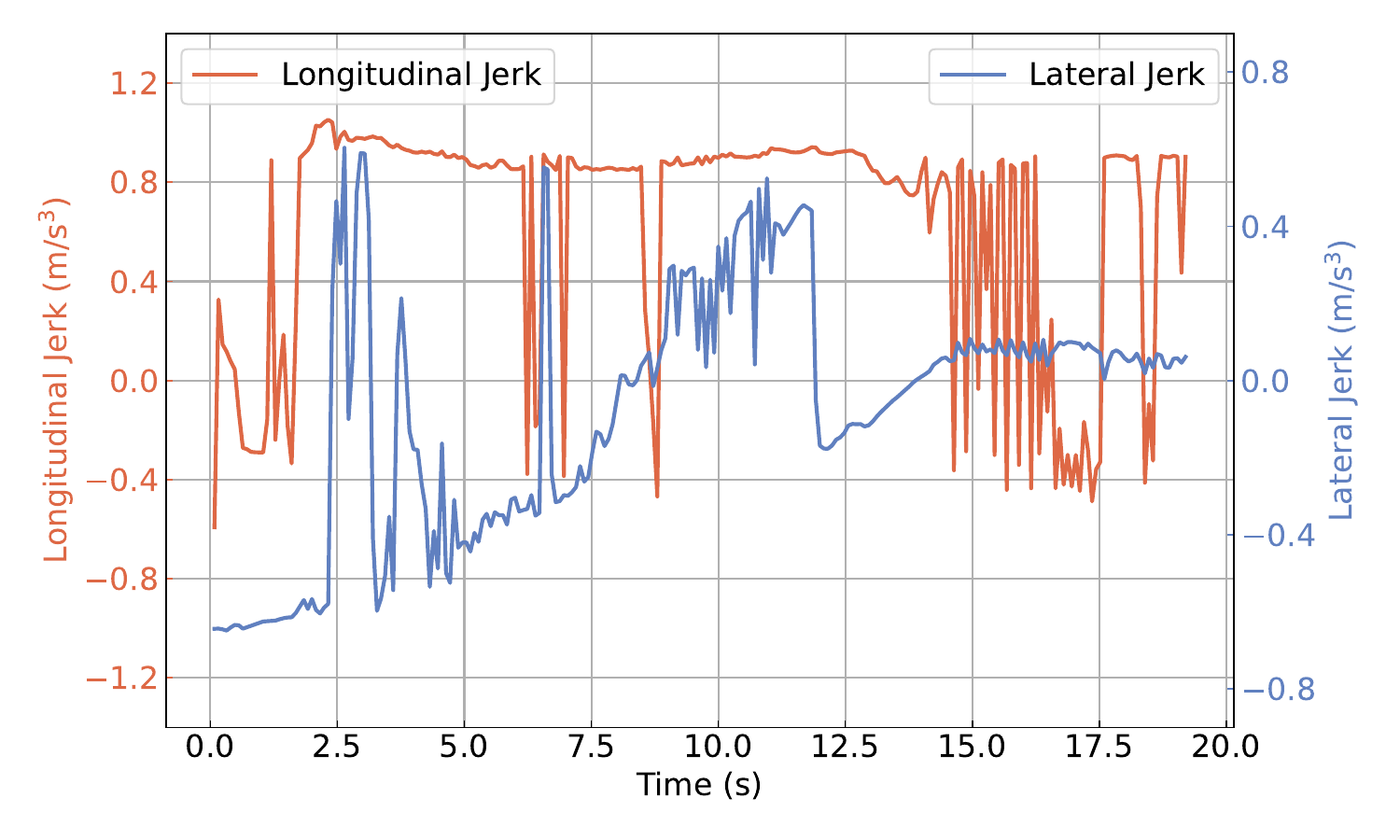}	\vspace{-2mm}
		\caption{Evolution of the longitudinal and lateral jerk profiles. The longitudinal and lateral jerk values lie within the desired boundary $[-0.9,0.9]$ and $[-0.6,0.6]$ for most of the time. }\vspace{-2mm}
		\label{fig:trajs_cruise_cutin_jerk}
	\end{figure}
  
   The trajectory and longitudinal velocity evolution are depicted in Fig.~\ref{fig:NGSIM_cruise_cutin_traj}, illustrating velocity changes at different positions during interactions with abrupt cut-in HVs. To obtain an intuitive understanding of this adjustment, Fig.~\ref{fig:NGSIM_cruise_cutin_vel_acc} shows the corresponding velocity and acceleration profiles of two HVs and the EV. Notably, at the commencement of Phase 2 (around $1.2\,\text{s}$), the EV decelerates to reduce its longitudinal velocity, demonstrating a proactive response to the abrupt cut-in by HV2, which has a larger acceleration range. The EV adaptively adjusts its longitudinal velocity in sync with the front HV2's evolution, ensuring a safe following distance.  Additionally, in Fig.~\ref{fig:trajs_cruise_cutin_jerk}, it is observed that the longitudinal and lateral jerk values are consistently maintained within a suitably configured range for the majority of the time. This demonstrates that the EV can stably react to the sudden cut-in by HV when performing lane-changing behaviors.

    \begin{figure}[tp]
    \centering
      \includegraphics[scale=0.355]{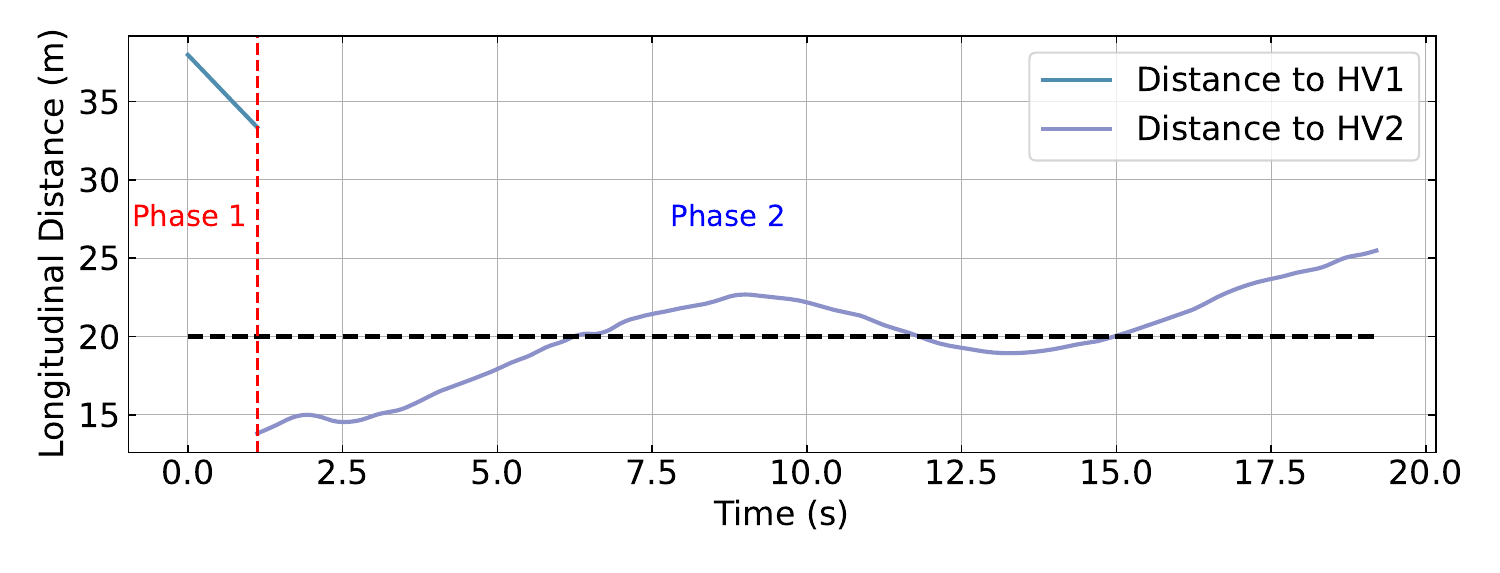}	\vspace{-2mm}
    \caption{Evolution of the headway when executing a lane-merging task under cruise scenario using the NGSIM dataset for surrounding HVs' motion. The dashed line denotes the safety barrier value during driving}\vspace{-2mm}
    \label{fig:trajs_cruise_cutin_headway}
\end{figure}
  
   Figure~\ref{fig:trajs_cruise_cutin_headway} visualizes the computed longitudinal distance to the nearest leading HV in two phases. Notably, the distance between the EV and HV2 violates the safety barrier value at the beginning and at around $12 \,\text{s}$. The initial violation stems from the sudden cut-in maneuver of the imperceptible HV2 at the beginning of Phase 2, while the subsequent breach is due to the front HV2's abrupt deceleration, surpassing the EV's maximum allowable deceleration limit, as illustrated in Fig.~\ref{fig:NGSIM_cruise_cutin_vel_acc}. After the first violation, the EV endeavors to decrease its speed to increase its following distance. However, the HV2 exhibits significant deceleration values from $1.4\,\text{s}$ to $2.3\,\text{s}$, surpassing the maximum allowable deceleration of the EV,  as depicted in Fig.~\ref{fig:NGSIM_cruise_cutin_vel_acc}. Consequently, the following distance experiences a slight decrease from $1.6 \,\text{s}$ to $2.5 \,\text{s}$. After that, the following distance asymptotically converges to the desired value $20 \,\text{m}$.
    The second violation is attributed to the severe deceleration fluctuation of the front HV2, leading to a decrease in the following distance. Following this violation, the EV stably adjusts its speed to achieve the desired following distance. These findings demonstrated our algorithm's robustness and safety recovery capabilities in response to unexpected safety barrier violations.
 
   In this lane-merging conflict scenario, where sudden maneuvers by HVs caused safety barrier violations, our algorithm demonstrates real-time adaptability by promptly adjusting the EV's speed to proactively address these safety issues. Despite transient decreases in the safety barrier value, the algorithm dynamically stabilized the situation, emphasizing its commitment to continuous safety improvement. This underscores the algorithm's robustness and safety recovery capabilities in addressing unexpected scenarios, reinforcing its reliability in enhancing safety within complex and dynamic driving environments.

    \section{Discussions}
	\label{sec:discussion}   
 \subsection{Real-Time Performance} 
    To evaluate the computational efficiency of our BPHTO framework, we manipulate the number of free-end homotopic trajectories $N_c$ and the nearest $M$ HVs considered in cruise tasks where the setting is the same as Section~\ref{sussubsec:dynamic_cruise}. 
    This allows us to analyze the computational time required across various simulation setups.
    \begin{figure}[tp]
    \centering
        \subfigure[]{
            \label{fig:num_traj} \hspace{-2mm}
        \includegraphics[scale=0.18]{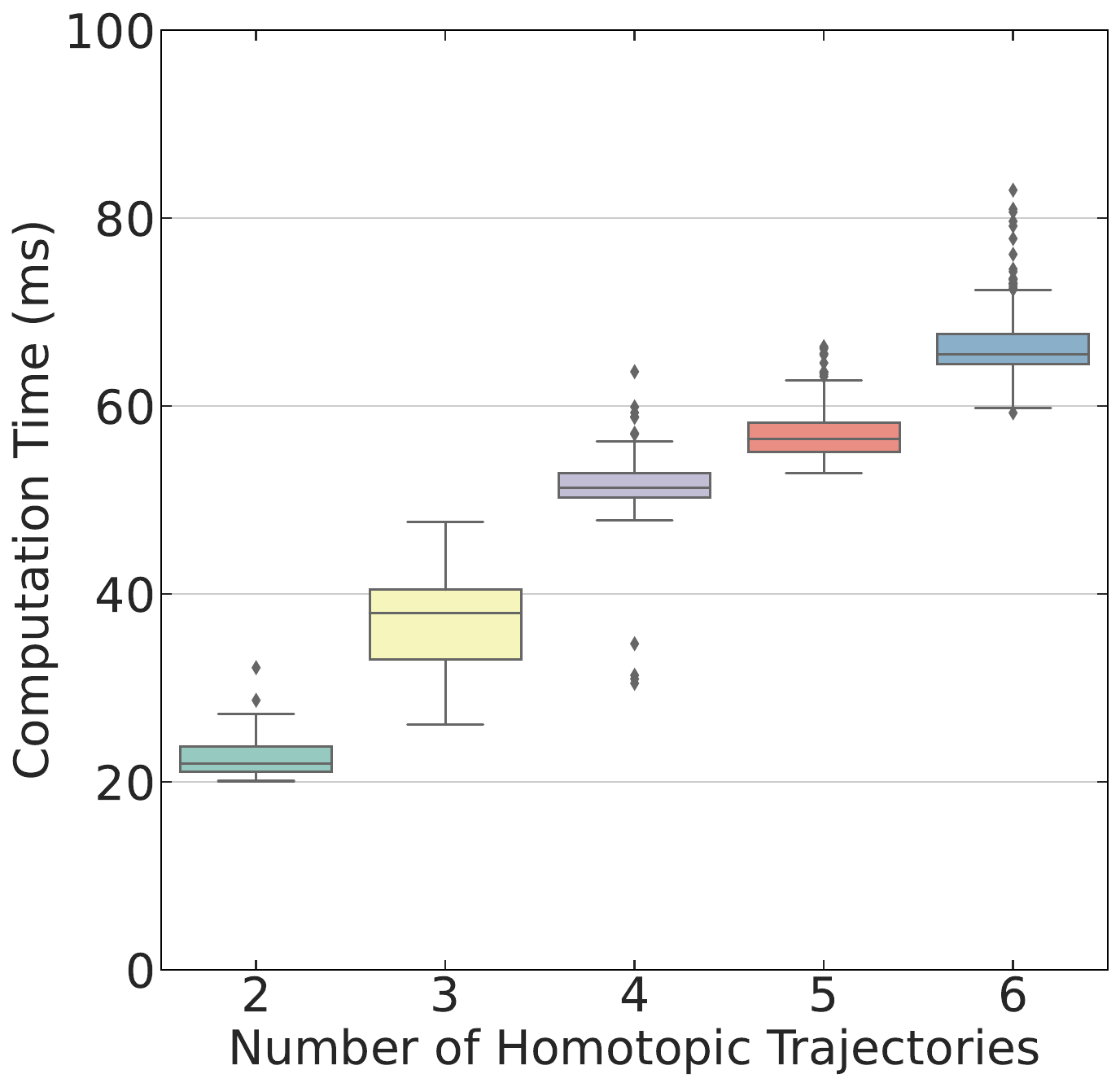}}\hspace{-2mm} 
        \subfigure[]{
            \label{fig:num_obs}
        \includegraphics[scale=0.18]{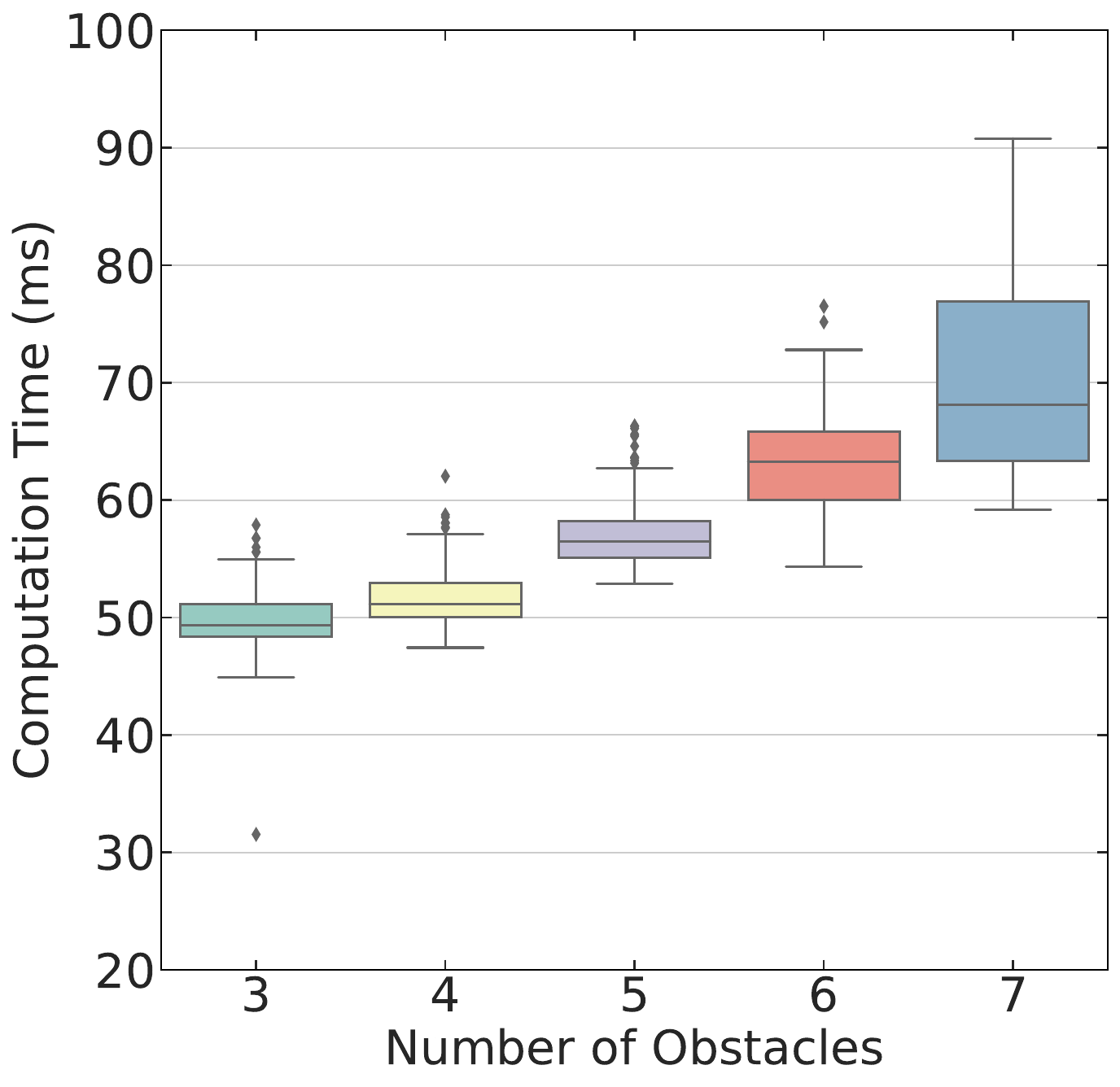}}\hspace{-2mm}
    \vspace{-1mm}
    \caption{Statistical results of the computation time per cycle with planning horizon $N = 50$. (a) Different
     number of homotopic trajectories $N_c$. (b) Different numbers of nearest considered HVs with five homotopic trajectories.}	
    \label{fig:diss_computational_time}
    \vspace{-1mm}
\end{figure}	  

    As depicted in Fig.~\ref{fig:num_traj}, a linear increase is observed in the average optimization time concerning the number of homotopic candidate trajectories. Given our current practice of employing a single thread for optimizing multiple trajectories,  one can facilitate computational efficiency through the application of multi-threading techniques in engineering applications. Additionally, we can notice that the average computation time of BPHTO is less than 100\,\text{ms} with the prediction length $N = 50$, as illustrated in Fig.~\ref{fig:num_obs}.
    These results indicate that our BPHTO algorithm attains real-time performance while handling different levels of density of obstacles. This efficiency is a key factor in ensuring real-time performance of our framework in dynamic and evolving traffic environments.
     \subsection{Driving Consistency} 
 
            \begin{figure}[tp]
    \centering
      \includegraphics[scale=0.405]{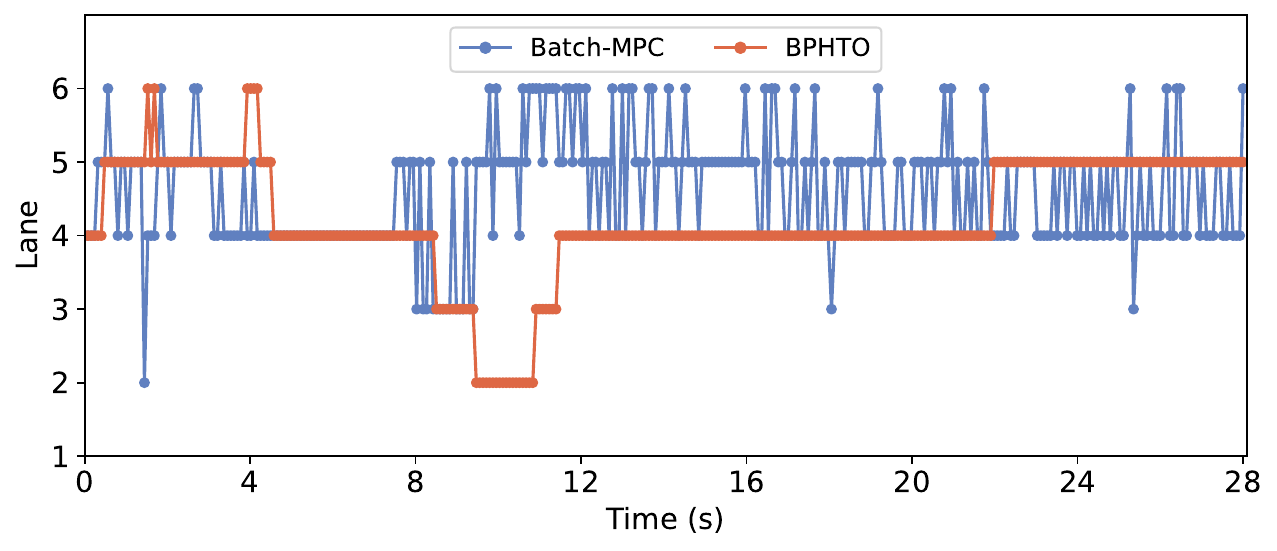}	\vspace{-2mm}
    \caption{Evolution of the target lane when executing an adaptive cruise under dense traffic using NGSIM dataset for surrounding HVs' motion.}\vspace{-2mm}
    \label{fig:NGSIM_Cruise_lane}
\end{figure}

To gain a more intuitive interpretation of driving consistency, 
the decision-making process illustrated in Fig.~\ref{fig:NGSIM_Cruise_lane} delineates the evolution of the target lane during the cruise in Section~\ref{sussubsec:dynamic_cruise}. Notably, Batch-MPC exhibits a tendency to frequently switch between different driving lanes in various directions, with some instances involving shifts across more than one lane. A prominent example is observed at $1.7\,\text{s}$, where Batch-MPC switches the target driving lane from lane 5 to lane 2. In contrast, BPHTO showcases a more consistent driving behavior, with no abrupt lane changes detected between two consecutive decision-making instants for the majority of the duration. The singular exception occurs at $1.9\,\text{s}$, where BPHTO undergoes a lane change between two adjacent lanes: lane 5 and lane 6.  
This observation underscores the enhanced driving consistency achieved by BPHTO. 
	\section{Conclusions}
	\label{sec:con}  
    This paper presents an integrated decision-making and planning scheme with the BPHTO algorithm for real-time safety-critical autonomous driving. The framework leverages BF and reachability analysis to enhance safety interactions and driving stability. BPHTO is then designed to optimize multiple behavior-oriented nominal trajectories concurrently in real time through the over-relaxed ADMM algorithm. The effectiveness of BPHTO is verified through various cruise control driving tasks, utilizing both real-world and synthetic datasets compared to baseline approaches, with the attendant outcome of improved task accuracy, driving stability, and consistency. Notably, BPHTO demonstrates robust safety recovery capabilities after abrupt interventions by HVs and the ability to navigate cluttered road construction areas. The real-time performance and driving consistency of the proposed BPHTO are thoroughly discussed in the context of cruise control tasks. As part of our future work, we plan to extend the algorithm to accommodate perception uncertainties in urban environments to achieve safe autonomous driving. 
    
{\appendix[Derivation of Longitudinal Distance Changes]
In Case 1,
the maximum longitudinal acceleration \(a_{x,\max}\) is not reached while the desired longitudinal velocity $v_d$ is reached. Then, the time intervals of the acceleration, deceleration, and constant segments can be computed as:
 \begin{subequations}  
     \label{eq:case1_time_seg} 
     \begin{align}  
       t_a = \frac{a_1-a_{x,0}}{j_{x,\max}},  \\
       t_d = \frac{a_1}{-j_{x,\min}} =  \frac{a_1}{j_{x,\max}}, \\
       t_c = T- t_a -t_d,
      \end{align}
 \end{subequations}
where the temporary maximum acceleration $a_1$ is to be computed.
Additionally, we can get the velocity change as the integral of acceleration with respect to time as follows:
 \begin{equation} 
       \Delta V = v_\text{d} - v_{x,0}  = \frac{(2t_a+t_d)\times a_1 - (a_1-a_{x,0})\times t_a  }{2}.
     \label{eq:case1_delta_v}
   \end{equation}
 
Substituting the time segments \eqref{eq:case1_time_seg} into \eqref{eq:case1_delta_v}, we can get the value of $a_1$ as follows:
 \begin{equation} 
       a_1  = \sqrt{ \frac{2\Delta V j_{x,\max}+a^2_{x,0}}{2}}.
     \label{eq:case1_delta_v}
   \end{equation}
 
As a result, the corresponding longitudinal distance changes in Case 1 can be derived as:
\begin{alignat}{2} 
    \delta p_x = v_{x,0}t_a &+ \frac{a_{x,0} +a_1}{4} t^2_a + \left(v_{x,0} + \frac{a_{x,0}+a_1}{2} t_a\right)t_d \nonumber \\
    &\quad+ \frac{a_1}{4} t^2_d + v_\text{d}t_c.\label{eq:case1_delta_lon_dis} 
 \end{alignat}

   In Case 2, the maximum longitudinal acceleration \(a_{x,\max}\) is reached while the maximum velocity needs to be computed. Then, the time intervals of the acceleration, constant, and deceleration segments can be analytically obtained as follows:
 \begin{subequations}\label{eq:case2_time_seg}
     \begin{align}  
       t_a = \frac{a_{x,\max}-a_{x,0}}{j_{x,\max}},  \\ 
       t_c = \frac{a_{x,\max}}{-j_{x,\min}} = \frac{a_{x,\max}}{ j_{x,\max}}, \\
          t_d = T - t_a- t_d .
       \end{align}
    \vspace{-\baselineskip}
\end{subequations}

Similarly, it gives
 \begin{equation} 
    \begin{aligned}
       \Delta V &= \frac{(a_{x,0}+a_{x,\max})t_a + 2a_{x,\max}t_c+ a_{x,\max}t_d }{2}  \\
       &= \frac{a_{x,0}t_a + a_{x,\max} (t_a + 2t_c + t_d) }{2}.
     \end{aligned}
     \label{eq:case2_delta_v}
  \end{equation} 
  
Then, the corresponding longitudinal distance changes in Case 2 can be derived as:
\begin{equation} 
\begin{aligned}
    \delta p_x &= v_{x,0}t_a + \frac{a_{x,0} +a_{x,\max}}{4} t^2_a \\
    &\quad+ \left(v_{x,0} + \frac{a_{x,0}+a_{x,\max}}{2} t_a\right)t_c + \frac{a_{x,\max}}{2}t^2_c \\
    &\quad + \left(v_{x,0} + \frac{a_{x,0}+a_{x,\max}}{2} t_a + a_{x,\max} t_c\right) t_d \\
    &\quad + \frac{a_{x,\max}}{4} t^2_d .
\end{aligned}
\label{eq:case2_delta_lon_dis}
  \end{equation} 
 
 }
 	\bibliographystyle{IEEEtran}
	\bibliography{egbib}
 
\end{document}